\documentclass[11pt]{article}

\usepackage{acl}

\usepackage{times}
\usepackage{latexsym}
\usepackage{amsmath}
\usepackage{amssymb}
\usepackage{algorithm}
\usepackage{algpseudocode}
\usepackage{tabularx}
\usepackage{comment}

 \usepackage{booktabs}
 \usepackage{multirow}
 \usepackage{siunitx}
\usepackage[T1]{fontenc}

\usepackage[utf8]{inputenc}

\usepackage{microtype}

\usepackage{inconsolata}

\usepackage{graphicx}
\usepackage{booktabs}
\usepackage{subcaption}
\usepackage{multirow}
\usepackage{array}
\usepackage{caption} 
\usepackage[autolanguage]{numprint}
\usepackage{pgfplots}
\usepackage{pifont}
\usepackage{xcolor}
\usepackage{float}
\usepackage[table]{xcolor}
\usepackage{longtable}
\usepackage{geometry}
\renewcommand{\arraystretch}{1.2}
\usepackage{supertabular}

\usepackage{pgfplots}
\usepackage{pgfplotstable}
\usepackage{subcaption}
\pgfplotsset{compat=1.18}
\usepgfplotslibrary{groupplots}
\usepackage{listings}

\DeclareCaptionType{listing}[Figure][List of Figures]

\definecolor{codebg}{RGB}{248,248,248}
\definecolor{codecomment}{RGB}{106,153,85}
\definecolor{codekeyword}{RGB}{0,0,180}
\definecolor{codestring}{RGB}{163,21,21}
\definecolor{codenumber}{RGB}{120,120,120}

\lstdefinestyle{pythonstyle_old}{
    language=Python,
    backgroundcolor=\color{codebg},
    basicstyle=\ttfamily\small,
    keywordstyle=\color{codekeyword}\bfseries,
    commentstyle=\color{codecomment}\itshape,
    stringstyle=\color{codestring},
    numberstyle=\tiny\color{codenumber},
    numbers=left,
    stepnumber=1,
    numbersep=10pt,
    showspaces=false,
    showstringspaces=false,
    showtabs=false,
    frame=single,
    rulecolor=\color{black!20},
    tabsize=4,
    captionpos=b,
    breaklines=true,
    breakatwhitespace=false,
    keepspaces=true,
    columns=fullflexible,
    xleftmargin=15pt,
    framexleftmargin=12pt,
    aboveskip=10pt,
    belowskip=10pt
}

\definecolor{codebg}{RGB}{242,242,242}         
\definecolor{codekeyword}{RGB}{44,88,201}      
\definecolor{codeclass}{RGB}{180,75,20}        
\definecolor{codestring}{RGB}{180,20,20}        
\definecolor{codecomment}{RGB}{0,135,45}    
\definecolor{codetype}{RGB}{110,60,180}        
\definecolor{codenumber}{RGB}{160,160,160}     
\definecolor{codeframe}{RGB}{220,220,220}      
\definecolor{argcolor}{RGB}{90,90,220}
\definecolor{classname}{RGB}{180,75,20}

\lstdefinelanguage{CustomPython}{
    language=Python,
    morekeywords={@dataclass},
    emph={str,List},
    emphstyle=\color{codetype},
}

\lstdefinestyle{pythonstyle}{
    language=CustomPython,
    backgroundcolor=\color{codebg},
    basicstyle=\ttfamily\small,
    keywordstyle=\color{codekeyword}\bfseries,
    commentstyle=\color{codecomment}\itshape,
    stringstyle=\color{codestring},
    numberstyle=\tiny\color{codenumber},
    numbers=left,
    stepnumber=1,
    numbersep=10pt,
    showspaces=false,
    showstringspaces=false,
    showtabs=false,
    frame=single,
    rulecolor=\color{codeframe},
    tabsize=4,
    captionpos=b,
    breaklines=true,
    breakatwhitespace=false,
    keepspaces=true,
    columns=fullflexible,
    xleftmargin=15pt,
    framexleftmargin=12pt,
    aboveskip=10pt,
    belowskip=10pt,
}

\usepackage{caption}
\usepackage{longtable}
\usepackage{siunitx}

\DeclareMathOperator*{\argmax}{arg\,max}
\title{EAGER: Enhancing Generative Event Extraction via Reinforcement Learning with Verifiable Rewards }

\author{
  \textbf{Omar Adjali\textsuperscript{1}},
  \textbf{Siting Liang\textsuperscript{1,2}},
  \textbf{Omair
Shahzad Bhatti\textsuperscript{1}},
  \textbf{Daniel Sonntag\textsuperscript{1,2}}
\\
\textsuperscript{1}German Research Center for Artificial Intelligence (DFKI), Germany
\\
\textsuperscript{2}Carl von Ossietzky Universität Oldenburg, Germany
\\
\small{
\{omar.adjali, siting.liang, omair\_shahzad.bhatti, 
daniel.sonntag\}@dfki.de
}
}

\newcommand{\gain}[1]{\textcolor[rgb]{0.0,0.5,0.0}{\scriptsize +#1}}

\definecolor{impbase}{RGB}{220,232,250}   
\definecolor{degbase}{RGB}{250,235,214}   

\definecolor{errimp}{RGB}{220,245,220}
\definecolor{errdeg}{RGB}{245,220,220}

\definecolor{impLight}{RGB}{240,246,255}
\definecolor{impMed}{RGB}{222,235,250}
\definecolor{impStrong}{RGB}{198,220,245}

\definecolor{degLight}{RGB}{255,247,235}
\definecolor{degMed}{RGB}{250,235,214}
\definecolor{degStrong}{RGB}{244,214,182}

\newcommand{\impsmall}[1]{\cellcolor{impLight}#1}
\newcommand{\impmed}[1]{\cellcolor{impMed}#1}
\newcommand{\implarge}[1]{\cellcolor{impStrong}#1}

\newcommand{\degsmall}[1]{\cellcolor{degLight}#1}
\newcommand{\degmed}[1]{\cellcolor{degMed}#1}
\newcommand{\deglarge}[1]{\cellcolor{degStrong}#1}

\begin{document}
\maketitle

\begin{abstract}
End-to-end event extraction remains challenging for large language models as it requires simultaneous identification of event triggers, classification of event types, and extraction of schema-grounded argument spans. We present EAGER, a reinforcement learning framework for generative event extraction that combines fine-grained verifiable rewards with Schema-Contrastive Advantage Estimation to alleviate advantage collapse under sparse binary rewards. Our reward design explicitly targets structural validity, extraction accuracy, groundedness, coverage, over-generation, and span precision. Experiments across seven benchmark datasets show that EAGER consistently outperforms prompting, supervised fine-tuning, and prior reinforcement learning baselines, achieving a substantial improvement over the strongest prior method. Results demonstrate that task-aligned verifiable rewards and contrastive advantage estimation substantially improve structured extraction.
\end{abstract}

\section{Introduction}
\begin{figure}[t]
    \centering
    \includegraphics[width=\linewidth]{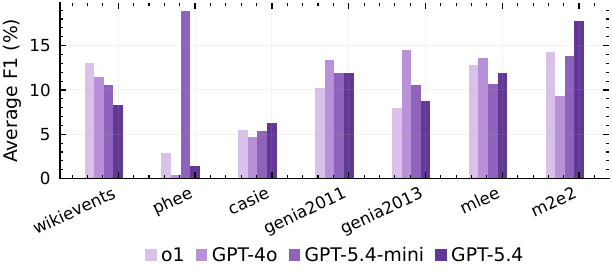}
    \caption{Average-F1 performance of frontier LLMs on end-to-end event extraction 
    across 7 datasets and domains.}
    \label{fig:approach_overview}
\vspace{-6mm}   
\label{fig:gpt_performance}
\end{figure}

Event extraction (EE) is a fundamental and challenging Information Extraction (IE) task that 
aims to identify event triggers, classify event types, and assign semantic roles to 
extracted arguments from unstructured text. With the advent of large language models, a 
growing body of work \cite{li-etal-2023-intra,lu-etal-2021-text2event,hsu2022degree,
ma2022prompt,wang2023instructuie,ren2023retrieve} has reformulated EE as a generative 
problem, leveraging sequence-to-sequence models to directly produce structured event representations under flexible annotation schemes. More recent approaches have further adopted LLMs through prompt engineering and chain-of-thought reasoning 
\cite{cai2024improving,gao2023exploring,hong2024towards,ma2024star}, aiming to reduce the computational overhead and domain-specific overfitting associated with fully supervised training. Nevertheless, as depicted in Figure~\ref{fig:gpt_performance}, even frontier general-purpose LLMs still fall short of achieving competitive performance on end-to-end 
EE benchmarks. A complementary line of work \cite{sainz2024gollie,
srivastava-etal-2025-instruction} has highlighted the critical role of augmenting training 
data with structured annotation guidelines, enabling instruction-tuned LLMs to better 
adhere to predefined event schemas and improve extraction fidelity. Despite these advances, 
significant performance gaps remain across diverse event types and domains.

\noindent To address these limitations, \citet{gao2024eventrl} proposed EventRL, which 
enhances LLM-based event extraction via outcome-supervised reinforcement learning (RL), 
optimizing the model based on the quality of final extracted event structures rather than 
relying solely on token-level supervision. This is further motivated by broader findings 
in the literature: comparative studies of supervised fine-tuning (SFT) versus RL-based 
fine-tuning \cite{huan2025does,chu2025sft} consistently show that RL-tuned models exhibit 
stronger cross-domain generalization and greater adaptability, whereas SFT-trained models are more susceptible to catastrophic forgetting, often degrading previously acquired 
general capabilities.

\noindent However, a key limitation in \citet{gao2024eventrl} and of existing training strategies 
more broadly, is that they provide only \textit{coarse-grained} supervision over extraction 
quality. Supervised fine-tuning optimizes next-token likelihood rather than extraction 
quality directly, while standard RL objectives typically rely on format validity and 
task-level accuracy alone. In practice, these signals are often too sparse to distinguish 
among distinct classes of extraction errors. Consequently, outcome-level reward signals 
struggle to explicitly target the issues inherent to generative EE such as hallucinated 
triggers, unsupported arguments, over-generation, and incomplete event coverage. A further 
structural limitation arises in group-based policy optimization: when all sampled 
completions within an optimization group are conditioned on the same prompt, including 
an identical set of negative schemas outputs tend to be homogeneous, producing near-zero reward variance and uninformative gradient signal.

\noindent In this work, we present EAGER, a post-training framework that addresses both 
limitations through two complementary contributions. First, we introduce a task-aligned, 
decomposed reward framework that explicitly models distinct quality dimensions of event 
extraction: \textit{validity}, \textit{extraction accuracy}, \textit{groundedness}, 
\textit{over-generation control}, \textit{coverage}, and \textit{span precision}. By 
decomposing the reward signal along these axes, our framework provides fine-grained 
optimization guidance that better aligns RL with the constraints of generative EE. Second, we propose \textit{Schema-Contrastive Advantage 
Estimation} (SCAE), which alleviates advantage collapse by independently sampling 
distinct sets of negative schemas across completions within each optimization group. This 
structural diversity induces variability in event type discrimination difficulty, 
increasing intra-group reward variance and enabling more informative policy gradients 
throughout training.

\section{Related Work}
Large language models have increasingly been adapted to structured information extraction through instruction tuning and schema-guided prompting~\cite{jiao2023instruct,lu2023pivoine}. Prior work has explored instruction-based IE frameworks such as InstructUIE~\cite{wang2023instructuie}, annotation-guided prompting in GoLLIE~\cite{sainz2024gollie}, and large-scale instruction resources like IEPile~\cite{gui2024iepile} to improve schema adherence and cross-domain generalization. Related work has also reformulated IE as code generation, where structured outputs benefit from the syntactic constraints of programming languages, as explored in CodeIE~\cite{li2023codeie} and KnowCoder~\cite{li2024knowcoder}.
\subsection{Event Extraction with Large Language Models}

Traditional event extraction approaches encompass both pipeline-based and joint modeling paradigms.~\citet{huang2024textee} conducted a comprehensive reevaluation of major EE paradigms on a standardized benchmark and found that current LLMs still fall short on several core EE subtasks, underscoring the gap between general instruction-following ability and task-specific extraction precision.~\citet{chen2024large} explored leveraging LLMs as automated annotators to generate high-quality event labels, effectively bootstrapping downstream model training while reducing human annotation costs. More recent work has examined how instruction tuning and annotation guidelines shape EE performance. \citet{srivastava-etal-2025-instruction} demonstrated that providing detailed textual descriptions of event types and argument roles during instruction tuning yields improved generalization to low-frequency and cross-schema event types. These studies highlight the persistent challenges of schema adherence, output validity, and domain generalization in event extraction.

\subsection{Preference and Reinforcement Learning for Structured Extraction}

Beyond supervised instruction tuning, a growing body of work has explored aligning LLMs to structured extraction objectives through preference learning and reinforcement learning. EventRL~\cite{gao2024eventrl} introduced an RL-based framework with outcome-driven reward functions targeting event identification and argument extraction, achieving gains in structural fidelity and generalization to novel event types.
More recently,~\cite{adjali2026aligning} extended this line toward multi-objective alignment for event extraction, combining task-level, format, and retrieval rewards within a GRPO framework. This contrasts with a closely related research direction which focuses on preference optimization tailored to IE tasks. In particular, ADELIE~\cite{qi2024adelie} combines supervised fine-tuning on curated instruction datasets with Direct Preference Optimization (DPO) using IE-specific comparison pairs, explicitly targeting schema adherence, argument completeness, and format validity. This approach achieves strong performance across closed, open, and on-demand IE settings while preserving general reasoning capability, suggesting that decomposing alignment objectives along task-relevant dimensions yields more reliable structured outputs.
These work show that RL-based approaches to event extraction remain notably underexplored and suggest that fine-grained alignment via reward design and preference learning can substantially improve the reliability of LLMs for structured IE, complementing purely supervised approaches.

\section{Method}

\begin{figure}[ht]
    \centering
    \includegraphics[width=\linewidth]{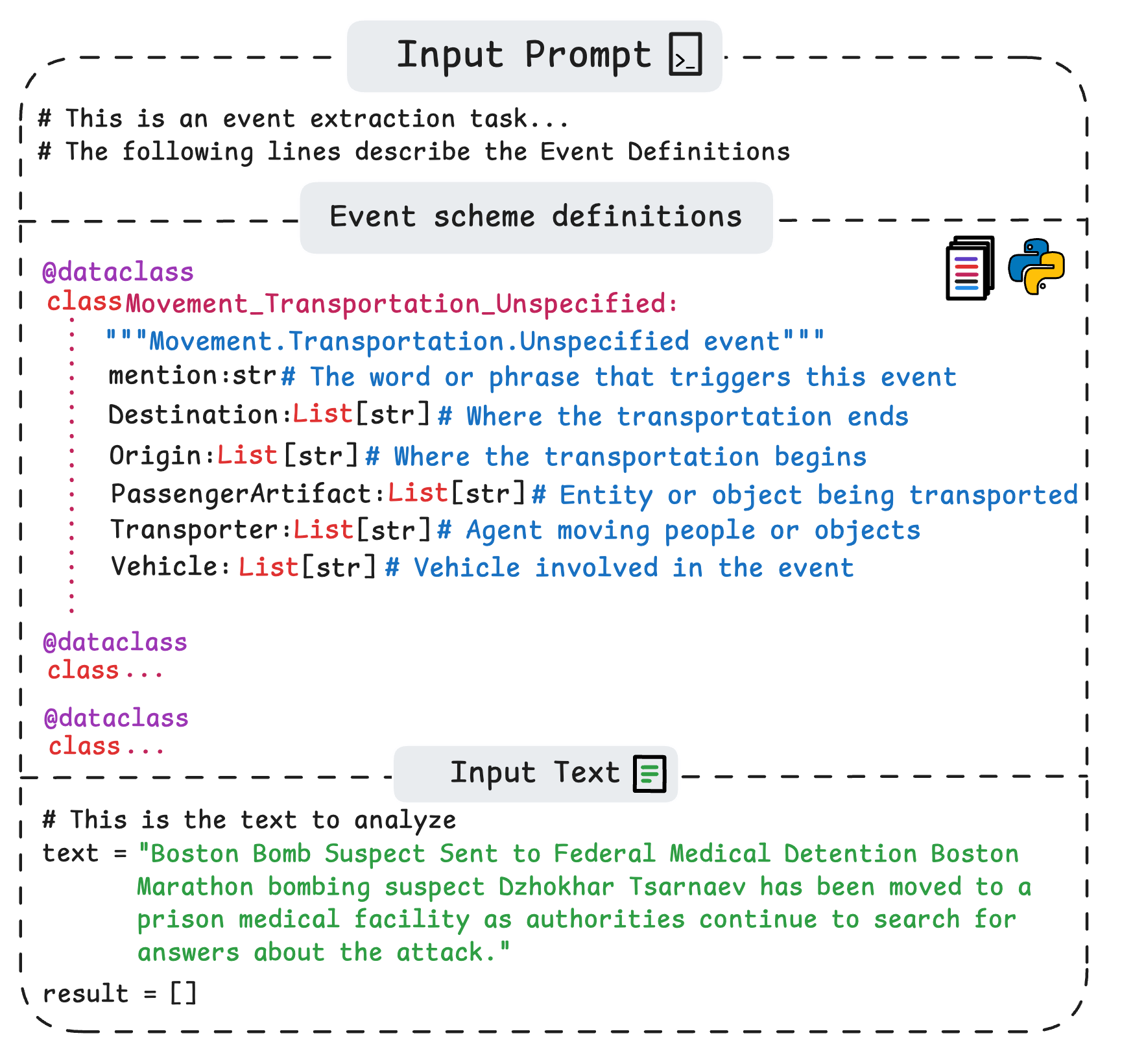}
    \caption{Structure of a prompt for end-to-end Event Extraction, comprising a task instruction a set of event schemas and the input text.}
    \label{fig:prompt}
\vspace{-4mm}  
\label{fig:prompt_structure}
\end{figure}

\subsection{Task Formulation}

Event extraction is a structured prediction task that encompasses four interdependent subtasks.
\textbf{Trigger Identification (TI)} detects event-denoting spans within an input text $X$.
\textbf{Trigger Classification (TC)} assigns a semantic event type to each identified trigger.
\textbf{Argument Identification (AI)} locates textual spans that participate as event arguments.
\textbf{Argument Classification (AC)} maps each identified argument span to a predefined semantic role. Formally, let $\mathcal{E} = \{E_i\}_{i=1}^{n}$ denote a predefined event schema, where each schema $E_i$ specifies a set of permissible argument roles. Given an input text $X$, the objective is to produce a structured event representation:
\begin{equation}
    Y = \{(t,\, r,\, a)\},
\end{equation}
where $t \in X$ denotes a trigger span, $r$ an argument role defined in $\mathcal{E}$, and $a \in X$ an argument span. EE thus learns a mapping $f_\theta : X \rightarrow Y$ subject to the schema constraints imposed by $\mathcal{E}$.

\subsection{Code-Based Input and Output Representations}
\label{sec:code_repr}

Following GoLLIE~\cite{sainz2024gollie}, we cast EE as Python code generation (Fig.~\ref{fig:prompt}). This leverages LLMs’ structural understanding of code~\cite{wang2023code4struct,li2023codeie} while providing a unified, less ambiguous representation for structured prediction~\cite{sainz2024gollie,srivastava-etal-2025-instruction}. Python syntax also guarantees well-formed outputs and simplifies parsing. Each event schema $E_i \in \mathcal{E}$ is defined as a \texttt{@dataclass} (Fig.~\ref{fig:event_python_class}), and extracted events are represented as class instances.

\subsection{Annotation Guideline Generation}
\label{sec:guidelines}

Although code-based schemas are compact and structured, they lack the semantic detail of annotation manuals. Prior work shows that LLM-generated guidelines can match manually written annotations for enriching event representations in instruction tuning~\cite{srivastava-etal-2025-instruction}. We therefore augment each Python class schema with natural-language guidelines generated via instruction-tuned prompting; details are provided in Appendix~\ref{appendix:Guidelines Annotation Augmentation}.

\subsection{Prompt Structure}
Each training instance is constructed as a structured prompt sequence as illustrated in Figure~\ref{fig:prompt_structure}:
\begin{equation}
    P = I \oplus E_e^{\text{G}} \oplus X,
\end{equation}
where $I$ denotes a natural language task instruction, $E_e^{\text{G}}$ is the gold event schema of type $e$ augmented with its automatically generated annotation guideline, and $X$ is the input text. This formulation encourages the model to jointly attend to schema constraints, including the set of permissible argument roles for $e$, and the contextual semantics conveyed by the annotation guidelines.

\noindent As shown in~\cite{gui2024iepile}, to further improve event type discrimination, we additionally sample a set of negative schemas, i.e., schemas corresponding to event types not present in $X$, and include them in the training prompt. See Figure~\ref{fig:example_input_prompt} for an example of an input prompt.

\subsection{Post-Training Framework}

Reinforcement learning with verifiable rewards (RLVR), powered by algorithms such as GRPO~\cite{shao2024deepseekmath} and DAPO~\cite{yu2026dapo}, has demonstrated strong effectiveness across a range of reasoning tasks ~\cite{guo2025deepseek}, however the application of RLVR to 
structured information extraction tasks  such as end-to-end event extraction, 
which requires simultaneously identifying event triggers, classifying event types, 
and extracting schema-grounded argument spans remains largely unexplored. 
To bridge this gap, we investigate DAPO 
~\cite{yu2026dapo} to enhance the reasoning ability of LLMs for EE, leveraging its verifiable, schema-based reward signal to guide structured output generation.

\noindent Given an input prompt $X$, the model samples a group of candidate outputs and optimizes the  DAPO objective

\begin{multline}
  \mathcal{J}_{\mathrm{DAPO}}(\theta)
  = \mathbb{E}\!\left[
    \frac{1}{\sum_{i}|o_i|}
    \sum_{i=1}^{G}\sum_{t=1}^{|o_i|}
    \min\!\Bigl(r_t^i\hat{A}_t^i,\right.\\
    \left.\operatorname{clip}(r_t^i,\,
    1{-}\epsilon_{\mathrm{low}},\,
    1{+}\epsilon_{\mathrm{high}})\,\hat{A}_t^i\Bigr)
  \right]
  \label{eq:dapo-objective}
\end{multline}
where $r_t^i(\theta) = {\pi_{\theta}(o_{i,t} \mid X, o_{i,<t})}/
{\pi_{\theta_{\mathrm{old}}}(o_{i,t} \mid X, o_{i,<t})}$ is the importance
sampling ratio and $\hat{A}_t^i$ is the group-normalized advantage.

\subsection{Reward-Decoupled Normalization}

Weighted reward sums can be dominated by high-variance components, suppressing weaker signals and requiring manual tuning. Following GDPO~\cite{liu2026gdpo}, we use Reward-Decoupled Normalization (RDN), normalizing each reward independently within the sampled group as $\tilde{R}_i^{(m)}=\frac{R_i^{(m)}-\mu^{(m)}}{\sigma^{(m)}+\epsilon}$, where $\mu^{(m)}$ and $\sigma^{(m)}$ are the group mean and standard deviation for reward $m$. The final reward is $\hat{R}_i=\sum_{m=1}^{M}\tilde{R}_i^{(m)}$. This avoids manual weighting, balances reward contributions, and alleviates reward hacking by preventing any single component from dominating the optimization.

\subsection{SCAE: Schema-Contrastive Advantage Estimation}

Similar to GRPO, DAPO discards the value network of PPO~\cite{schulman2017proximal} by computing advantages directly from group-level outcome rewards. 
For each input text $X$ and its gold event annotation $Y$, we
samples a group of $G$ outputs $\{o_i\}_{i=1}^{G}$ from the old policy
$\pi_{\theta_{\text{old}}}$, assigns binary outcome rewards $\{R_i\}_{i=1}^{G}$, and
estimates the per-token advantage as the group-normalized reward:

\begin{equation}
  \hat{A}_t^i =
    \frac{R_i - \mathrm{mean}\!\left(\{R_i\}_{i=1}^{G}\right)}
         {\mathrm{std}\!\left(\{R_i\}_{i=1}^{G}\right)},
  \label{eq:grpo-advantage}
\end{equation}
However, a critical issue arises when all sampled completions within a group receive identical rewards 
leading to near-zero policy gradients and stalled learning \cite{zhang2025scaf,yu2026dapo,he2026avspo}. In standard DAPO sampling setting, all $G$ completions are conditioned on the same prompt $P = I \oplus E_e^{\text{G}} \oplus X$, including an identical set of negative schemas. Since the negative schemas strongly constrain the model's event type discrimination signal, sampled outputs tend to exhibit low diversity, such that $\mathrm{Var}(\{R_i\}_{i=1}^{G} \mid P) \approx 0$. Combined with sparse binary rewards, this frequently produces homogeneous reward groups with uninformative gradient signal.

\noindent We propose SCAE, a schema-contrastive formulation of advantage estimation that structurally injects reward variance by varying the set of negative schemas across completions within the same optimization group, rather than holding the full prompt fixed.
During training, each prompt $P$ includes not only the gold schema $E_e^{\text{G}}$ for the target event type $e$, but also a set of $K$ negative schemas sampled from $\mathcal{E} \setminus \{e\}$. Formally, let $\mathcal{N} = \mathcal{E} \setminus \{e\}$ denote the pool of available negative schemas. Rather than fixing a single negative set across all group members, we construct a \emph{schema-contrastive group} by independently sampling a distinct subset $\mathcal{S}_i \subset \mathcal{N}$, $|\mathcal{S}_i| = K$, for each completion $i$, yielding group-specific prompts:
\begin{equation}
    P_i = I \oplus E_e^{\text{G}} \oplus \mathcal{S}_i \oplus X, \quad \mathcal{S}_i \overset{\text{i.i.d.}}{\sim} \binom{\mathcal{N}}{K}
\end{equation}
where $\mathcal{S}_i \neq \mathcal{S}_j$ for $i \neq j$ with high probability when $|\mathcal{N}| \gg K$. The contrastive group is then defined as:
\begin{equation}
    \mathcal{G}_e = \left\{ (P_i,\, \hat{o}_i) \right\}_{i=1}^{G}, \quad \hat{o}_i \sim \pi_\theta(\cdot \mid P_i)
\end{equation}
where each sampled completion $\hat{o}_i$ receives an independent reward $R_i$ computed against the gold annotation $Y$. By varying the negative schema context across group members, different completions are exposed to different distractor event types, inducing variability in the difficulty of event type discrimination and thereby increasing intra-group reward variance:
\begin{equation}
    \mathrm{Var}\!\left(\{R_i\}_{i=1}^{G}\right) \gg \mathrm{Var}\!\left(\{R_i\}_{i=1}^{G} \mid P\right)
\end{equation}

Advantages are then estimated by normalizing rewards within each schema-contrastive group and substituted into the DAPO objective (Eq.~\ref{eq:dapo-objective}). (See Algorithm~\ref{alg:scae_alg} for the detailed DAPO with SCAE pseudo-code.)

\begin{figure}[ht]
    \centering
    \includegraphics[width=\linewidth]{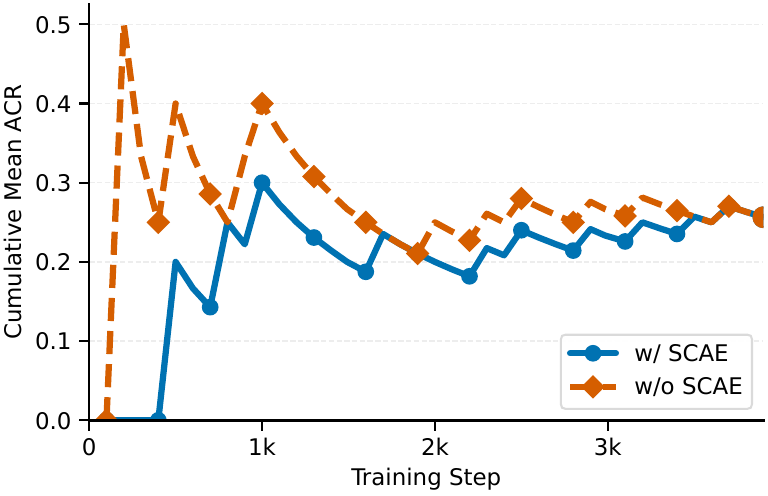}
    \caption{Cumulative Mean ACR during training}
    \label{fig:acr}
\vspace{-3mm}     
\end{figure}

To quantify the effectiveness of our schema-contrastive group construction, we compute the Advantage Collapse Rate (ACR)~\cite{he2026avspo}, which measures the proportion of training groups exhibiting near-zero reward variance:
\begin{equation}
    \mathrm{ACR} = \frac{1}{N}\sum_{j=1}^{N} \mathbb{I}\!\left(\sigma_{\mathcal{R}_j} < \tau\right),
\end{equation}
where $\sigma_{\mathcal{R}_j}$ is the reward standard deviation within group $j$ and $\tau$ is a small numerical threshold ($\tau \approx 10^{-6}$). An ACR close to 0 indicates that most groups 
produce informative gradient signals, while ACR $\approx 1$ signals complete gradient 
stagnation. As shown in Figure~\ref{fig:acr}, training with our Schema-Contrastive 
Advantage Estimation (w/ SCAE) consistently achieves a substantially lower cumulative 
ACR compared to the standard fixed-prompt baseline (w/o SCAE), confirming that varying 
the negative schema set across group members effectively alleviates reward homogeneity 
and produces more informative optimization signals throughout training.


\subsection{Verifiable Rewards Modeling}

\begin{table*}[t]
\centering
\scriptsize
\setlength{\tabcolsep}{6pt}
\renewcommand{\arraystretch}{1.15}
\begin{tabularx}{\textwidth}{@{} l l X X @{}}
\toprule
\textbf{Reward Component} & \textbf{Notation} & \textbf{Formulation} & \textbf{Role in Training} \\
\midrule
Validity
    & $R_{\text{fmt}}$
    & Binary: 1 if output parses into the required structured representation, 0 otherwise
    & Filters malformed outputs before any extraction scoring \\[4pt]

Extraction Accuracy
    & $R_{\text{EE}}$
    & Macro-average F1 over trigger identification (TI), trigger classification (TC), argument identification (AI), argument classification (AC), AI$^{+}$, and AC$^{+}$
    & Captures canonical end-to-end extraction quality \\[4pt]

Groundedness
    & $R_{\text{grd}}$
    & Verifies that predicted triggers and arguments appear verbatim in the source input; averages span-level support scores
    & Suppresses hallucinated spans and improves factual consistency \\[4pt]

Over-generation
    & $R_{\text{ovr}}$
    & Penalizes predictions that exceed the gold event and argument counts
    & Reduces false positives and limits excessive output length \\[4pt]

Coverage
    & $R_{\text{cov}}$
    & Proportional recall over gold events and arguments, normalized to prevent inflation from over-prediction
    & Encourages completeness while balancing over-generation penalties \\[4pt]

Span Precision
    & $R_{\text{span}}$
    & Token-level Jaccard similarity with an additional penalty for oversized predicted spans
    & Rewards near-correct boundary predictions and discourages span inflation \\
\bottomrule
\end{tabularx}
\caption{%
    Summary of the proposed verifiable rewards used in our  RL framework for generative event extraction. Each reward targets a distinct aspect: structural invalidity, extraction inaccuracy, hallucination, over-generation, under-coverage, and boundary imprecision.
}
\label{tab:reward_summary}
\end{table*}

Rather than relying solely on outcome reward using aggregate F1 scores such in \cite{gao2024eventrl}, we define specialized rewards described in Table~\ref{tab:reward_summary} that target complementary aspects of model behavior, including output validity, extraction correctness, groundedness, prediction balance, and span precision. This modular design provides more interpretable and fine-grained optimization signals. See Appendix~\ref{appendix:Verifiable Rewards Formulation} for more formal reward definitions. 

\begin{table*}[t]
	\centering
	\scriptsize
    
	\setlength{\tabcolsep}{6pt}
	\renewcommand{\arraystretch}{1.15}
	\begin{tabularx}{\textwidth}{l X X X X X X X X @{}}
		\toprule
		\textbf{Model} & \textbf{WikiEvents} & \textbf{PHEE} & \textbf{CASIE} & \textbf{GENIA11} & \textbf{GENIA13} & \textbf{MLEE} & \textbf{M2E2} & \textbf{Avg.} \\
		\midrule

		\multicolumn{9}{l}{\textit{Few-shot Prompting}} \\
		\cmidrule(lr){1-9}
        o1~\cite{jaech2024openai} 
        & 13.08 & 2.86 & 5.49 & 10.27 & 7.95 & 12.80 & 14.24 & 9.52 \\
        
		GPT-4o~\cite{hurst2024gpt} 
        & 10.31 & 2.00 & 3.47 & 12.43 & 10.87 & 13.35 & 11.97 & 9.20 \\
        
        GPT-5.4-mini~\cite{singh2025openai} 
        & 10.58 & 18.86 & 5.38 & 11.95 & 10.56 & 10.70 & 13.86 & 11.70 \\
        
        GPT-5.4~\cite{singh2025openai} 
        & 8.30 & 1.42 & 6.22 & 11.94 & 8.74 & 11.90 & 17.74 & 9.47 \\
        
		GoLLIE-7B~\cite{sainz2024gollie} 
        & 13.65 & 30.44 & 7.47 & 12.58 & 9.69 & 8.49 & 32.14 & 16.35 \\
        
		GoLLIE-13B~\cite{sainz2024gollie} 
        & 13.97 & 30.08 & 7.02 & 10.33 & 9.56 & 11.46 & 38.96 & 17.34 \\
        
		GoLLIE-34B~\cite{sainz2024gollie} 
        & 11.75 & 27.42 & 7.43 & 11.26 & 8.02 & 9.58 & 32.80 & 15.47 \\

		\addlinespace[2pt]
		\multicolumn{9}{l}{\textit{Supervised fine-tuning}} \\
		\cmidrule(lr){1-9}
		GoLLIE-7B + SFT 
        & 13.31 & 30.33 & 7.62 & 12.31 & 10.34 & 8.95 & 32.70 & 16.51 \\
        
		GoLLIE-13B + SFT 
        & 13.78 & 30.08 & 7.02 & 10.33 & 9.56 & 11.46 & 38.96 & 17.31 \\
        
        \cite{srivastava-etal-2025-instruction}$^\dagger$$_{\textnormal{LLaMA31-8B}}$ 
        & 5.78 & 28.70 & 7.95 & 13.18 & 9.04 & 9.44 & 12.96 & 12.44 \\

		\addlinespace[2pt]
		\multicolumn{9}{l}{\textit{Reinforcement learning}} \\
		\cmidrule(lr){1-9}       

		EventRL~\cite{gao2024eventrl}$^\dagger$
        & 15.18 & 52.40 & 8.90 & 16.65 & 16.31 & 11.84 & 43.53 & 23.54\\
        \textnormal{ADELIE}$_{\textnormal{DPO}}$~\cite{qi2024adelie}$^\dagger$ & 18.20 & 52.61 & 10.90 & 20.91 & 19.50 & 16.89 & 40.49 & 25.64 \\

        \midrule
		\textbf{EAGER (ours)}
		& \textbf{21.31} & \textbf{60.36} & \textbf{14.66} & \textbf{26.32} & \textbf{23.39} & \textbf{23.42} & \textbf{46.09} & \textbf{30.79} \\

        \textit{Gain vs. best prior}
        & \gain{3.11} & \gain{7.75} & \gain{3.76} & \gain{5.41} & \gain{3.89} & \gain{6.53} & \gain{2.56} & \gain{5.15} \\
        
		\bottomrule
	\end{tabularx}    
	\caption{
	Mean F1 results for end-to-end event extraction on the test split of the seven benchmark datasets. \textbf{Avg.} denotes the average F1 performance across datasets and EE subtasks. Best results are shown in bold. Green values indicate absolute improvement over the strongest prior baseline. $^\dagger$ denotes literature methods we re-implemented.
	}	
\vspace{-6mm}  
\label{tab:ee_results}
\end{table*}

\begin{table}[t]
\centering
\scriptsize
\setlength{\tabcolsep}{1.8pt}
\renewcommand{\arraystretch}{0.88}

\caption{Ablation study on the contribution of each proposed individual reward to the global end-to-end event extraction performance. Blue shading denotes improvement relative to the baseline configuration $R_{\text{EE}} + R_{\text{FMT}}$, while amber shading denotes degradation. Darker shading indicates larger absolute change. Bold indicates the best score per dataset. This color scheme logic applies to the subsequent tables.}
\resizebox{\columnwidth}{!}{
\begin{tabular}{@{}lcccccccc@{}}
\toprule
Setting/Rewards  & Wiki & PHEE & CASIE & G11 & G13 & MLEE & M2E2 & Avg \\
\midrule

ALL Rewards
& \implarge{21.31}
& \implarge{\textbf{60.36}}
& \degmed{14.66}
& \impsmall{26.32}
& \degsmall{23.39}
& \impmed{\textbf{23.42}}
& \implarge{\textbf{46.09}}
& \implarge{\textbf{30.79}} \\

\textit{$R_{\text{EE}}$} \textit{+ $R_{\text{FMT}}$}
& 20.12
& 53.66
& 15.70
& 26.11
& 24.10
& 21.62
& 30.22
& 27.36 \\

\hspace{3mm}\textit{+ $R_{\text{grd}}$}
& \impmed{22.40}
& \impmed{57.74}
& \degsmall{14.94}
& \impsmall{\textbf{26.83}}
& \degmed{22.48}
& \degmed{20.02}
& \implarge{36.72}
& \impmed{28.73} \\

\hspace{3mm}\textit{+ $R_{\text{ovr}}$}
& \impmed{\textbf{23.04}}
& \impmed{58.27}
& \degmed{12.96}
& \impsmall{26.54}
& \degmed{22.14}
& \degmed{19.33}
& \implarge{39.20}
& \impmed{28.78} \\

\hspace{3mm}\textit{+ $R_{\text{cov}}$}
& \impmed{21.38}
& \impmed{58.01}
& \impsmall{\textbf{16.12}}
& \impsmall{26.34}
& \impsmall{24.22}
& \degsmall{21.53}
& \impmed{31.52}
& \impmed{28.45} \\

\hspace{3mm}\textit{+ $R_{\text{span}}$}
& \impmed{21.46}
& \implarge{58.97}
& \impsmall{15.91}
& \impsmall{26.63}
& \impsmall{\textbf{24.27}}
& \degsmall{21.40}
& \impmed{34.26}
& \impmed{28.99} \\

\bottomrule
\end{tabular}
}
\label{tab:reward_ablation}
\vspace{-5mm}
\end{table}

\begin{table}[t]
\centering
\scriptsize
\setlength{\tabcolsep}{1.8pt}
\renewcommand{\arraystretch}{0.88}

\caption{Impact of augmenting event schemes with annotation guidelines.}
\resizebox{\columnwidth}{!}{
\begin{tabular}{@{}lcccccccc@{}}
\toprule
Setting & Wiki & PHEE & CASIE & G11 & G13 & MLEE & M2E2 & Avg \\
\midrule

GoLLIE-7B
& 13.65
& 30.44
& 7.47
& 12.58
& 9.69
& 8.49
& 32.14
& 16.35 \\

\hspace{3mm}\textit{w/o A.G.}
& \degmed{9.14}
& \deglarge{7.67}
& \degmed{3.13}
& \deglarge{1.64}
& \deglarge{1.17}
& \deglarge{2.41}
& \deglarge{15.54}
& \deglarge{5.81} \\

EAGER
& \textbf{21.31}
& \textbf{60.36}
& \textbf{14.66}
& \textbf{26.32}
& \textbf{23.39}
& \textbf{23.42}
& \textbf{46.09}
& \textbf{30.79} \\

\hspace{3mm}\textit{w/o A.G.}
& \degsmall{19.43}
& \deglarge{43.66}
& \degmed{8.34}
& \degmed{17.32}
& \degmed{16.96}
& \deglarge{12.25}
& \degsmall{40.33}
& \degmed{22.61} \\

\bottomrule
\end{tabular}
}
\label{tab:A.G_ablation}
\vspace{-3mm}
\end{table}

\section{Experimental Setup}

\paragraph{Baselines}
We compare EAGER against representative methods spanning prompting, supervised fine-tuning, and reinforcement learning paradigms for end-to-end event extraction. We include proprietary large language models evaluated under few-shot prompting, including o1~\cite{jaech2024openai}, GPT-4o~\cite{hurst2024gpt}, GPT-5.4-mini, and GPT-5.4~\cite{singh2025openai}, to assess the effectiveness of general-purpose reasoning-oriented LLMs without task-specific adaptation. We additionally compare against GoLLIE~\cite{sainz2024gollie}, a code-oriented instruction framework for information extraction. To isolate the contribution of reinforcement learning beyond standard instruction tuning, we fine-tune GoLLIE-7B and GoLLIE-13B using SFT. We also compare against the annotation-guideline augmented instruction tuning framework of~\cite{srivastava-etal-2025-instruction}, which similarly enriches event schemas with LLM-generated task descriptions. Finally, we compare against prior RL-based event extraction methods, including ADELIE$_{\textnormal{DPO}}$~\cite{qi2024adelie}, which applies direct preference optimization for generative information extraction, and EventRL~\cite{gao2024eventrl}, which optimizes event extraction performance using reinforcement learning with outcome-based rewards. These baselines represent the closest prior approaches to post-training optimization for structured extraction. For fair comparison, all reproduced baselines marked with $^\dagger$ are evaluated using the same evaluation protocols and under a unified experimental setup.

\begin{table}[t]
\centering
\caption{Detailed results of event trigger and argument performance including TI, TC, AI, AC, AI+, and AC+. Bold indicates the best score within each dataset. Improvement/Degradations are highlighted relative to the baseline configuration $R_{\text{EE}} + R_{\text{fmt}}$.}
\label{tab:extended_results}
\scriptsize
\setlength{\tabcolsep}{3pt}
\renewcommand{\arraystretch}{0.92}
\begin{tabular}{llcccccc}
\toprule
Dataset & Method & TI & TC & AI & AC & AI+ & AC+ \\
\midrule

\multirow{6}{*}{WikiEvents}
& All
& \degsmall{38.92}
& \degmed{30.83}
& \impmed{17.28}
& \impmed{14.51}
& \impmed{14.32}
& \impmed{12.01} \\
& $R_{\text{EE}} + R_{\text{fmt}}$
& 39.53 & 32.37 & 15.33 & 12.28 & 11.73 & 9.47 \\
& \hspace{3mm}+$R_{\text{span}}$
& \impsmall{39.88}
& \impmed{35.31}
& \impmed{16.96}
& \impmed{14.10}
& \impsmall{12.17}
& \impsmall{10.37} \\
& \hspace{3mm}+$R_{\text{ovr}}$
& \degsmall{39.22}
& \impmed{\textbf{35.37}}
& \impmed{\textbf{18.22}}
& \implarge{\textbf{16.12}}
& \implarge{\textbf{15.25}}
& \implarge{\textbf{14.05}} \\
& \hspace{3mm}+$R_{\text{grd}}$
& \degsmall{38.74}
& \impmed{34.12}
& \impmed{17.92}
& \implarge{15.89}
& \implarge{14.66}
& \implarge{13.09} \\
& \hspace{3mm}+$R_{\text{cov}}$
& \impmed{\textbf{41.53}}
& \impmed{34.65}
& \impsmall{15.83}
& \impsmall{13.52}
& \impsmall{12.20}
& \impsmall{10.52} \\
\midrule

\multirow{6}{*}{PHEE}
& All
& \impmed{\textbf{69.35}}
& \impmed{\textbf{68.74}}
& \implarge{\textbf{68.65}}
& \implarge{\textbf{60.47}}
& \implarge{\textbf{50.31}}
& \implarge{\textbf{44.67}} \\
& $R_{\text{EE}} + R_{\text{fmt}}$
& 67.92 & 66.80 & 62.40 & 46.17 & 45.03 & 33.65 \\
& \hspace{3mm}+$R_{\text{span}}$
& \degsmall{67.89}
& \impsmall{66.87}
& \implarge{67.07}
& \implarge{60.32}
& \implarge{48.38}
& \implarge{43.29} \\
& \hspace{3mm}+$R_{\text{ovr}}$
& \degsmall{67.62}
& \degsmall{66.60}
& \impmed{65.97}
& \implarge{59.50}
& \impmed{47.32}
& \implarge{42.60} \\
& \hspace{3mm}+$R_{\text{grd}}$
& \impsmall{68.29}
& \impsmall{67.28}
& \impsmall{64.07}
& \implarge{58.24}
& \impsmall{46.40}
& \implarge{42.15} \\
& \hspace{3mm}+$R_{\text{cov}}$
& \degsmall{67.89}
& \degsmall{66.77}
& \implarge{67.07}
& \implarge{57.12}
& \implarge{48.19}
& \implarge{41.02} \\
\midrule

\multirow{6}{*}{CASIE}
& All
& \degmed{18.54}
& \degsmall{17.19}
& \degmed{21.44}
& \impsmall{17.83}
& \degsmall{7.02}
& \impsmall{5.94} \\
& $R_{\text{EE}} + R_{\text{fmt}}$
& 20.63 & \textbf{19.92} & \textbf{23.60} & 17.12 & 7.55 & 5.40 \\
& \hspace{3mm}+$R_{\text{span}}$
& \degsmall{20.45}
& \degsmall{19.60}
& \degmed{22.05}
& \impmed{\textbf{19.16}}
& \impsmall{7.58}
& \impsmall{\textbf{6.60}} \\
& \hspace{3mm}+$R_{\text{ovr}}$
& \degmed{17.46}
& \degmed{16.57}
& \deglarge{16.73}
& \degmed{14.83}
& \degmed{6.40}
& \impsmall{5.75} \\
& \hspace{3mm}+$R_{\text{grd}}$
& \impsmall{\textbf{20.92}}
& \degsmall{19.85}
& \deglarge{18.54}
& \degsmall{16.77}
& \degsmall{7.09}
& \impsmall{6.48} \\
& \hspace{3mm}+$R_{\text{cov}}$
& \degsmall{20.58}
& \degsmall{19.75}
& \degsmall{23.50}
& \impmed{19.14}
& \impsmall{\textbf{7.61}}
& \impsmall{6.12} \\
\midrule

\multirow{6}{*}{Genia11}
& All
& \impsmall{\textbf{43.93}}
& \impsmall{\textbf{39.18}}
& \degsmall{25.05}
& \impsmall{23.85}
& \degsmall{13.19}
& \impsmall{12.73} \\
& $R_{\text{EE}} + R_{\text{fmt}}$
& 43.46 & 38.83 & 25.16 & 23.10 & 13.55 & 12.57 \\
& \hspace{3mm}+$R_{\text{span}}$
& \degsmall{42.95}
& \degsmall{38.20}
& \impmed{\textbf{26.63}}
& \impmed{24.40}
& \impsmall{14.28}
& \impsmall{13.34} \\
& \hspace{3mm}+$R_{\text{ovr}}$
& \degmed{42.33}
& \degsmall{38.55}
& \impsmall{25.59}
& \impsmall{24.14}
& \impmed{14.70}
& \impmed{13.92} \\
& \hspace{3mm}+$R_{\text{grd}}$
& \degmed{42.31}
& \degmed{37.39}
& \impmed{26.60}
& \impmed{\textbf{25.01}}
& \impmed{\textbf{15.22}}
& \impmed{\textbf{14.46}} \\
& \hspace{3mm}+$R_{\text{cov}}$
& \impsmall{43.65}
& \impsmall{39.14}
& \impsmall{25.38}
& \impsmall{23.31}
& \impsmall{13.80}
& \impsmall{12.77} \\
\midrule

\multirow{6}{*}{Genia13}
& All
& \degmed{42.51}
& \degsmall{37.87}
& \impsmall{\textbf{20.80}}
& \impmed{\textbf{19.75}}
& \degmed{9.83}
& \degmed{9.59} \\
& $R_{\text{EE}} + R_{\text{fmt}}$
& 43.64 & 38.73 & 20.67 & 18.64 & 11.92 & 11.02 \\
& \hspace{3mm}+$R_{\text{span}}$
& \impsmall{\textbf{44.60}}
& \impsmall{39.23}
& \degsmall{20.41}
& \degsmall{18.29}
& \impsmall{\textbf{12.07}}
& \impsmall{\textbf{11.02}} \\
& \hspace{3mm}+$R_{\text{ovr}}$
& \degmed{40.80}
& \degmed{36.55}
& \degmed{17.99}
& \degsmall{16.25}
& \degsmall{11.10}
& \degsmall{10.16} \\
& \hspace{3mm}+$R_{\text{grd}}$
& \degmed{41.90}
& \degmed{37.10}
& \degsmall{18.41}
& \degsmall{16.60}
& \degmed{10.88}
& \degmed{9.98} \\
& \hspace{3mm}+$R_{\text{cov}}$
& \impsmall{44.13}
& \impsmall{\textbf{39.44}}
& \degsmall{20.44}
& \degsmall{18.59}
& \degsmall{11.82}
& \degsmall{10.90} \\
\midrule

\multirow{6}{*}{MLEE}
& All
& \impmed{\textbf{46.55}}
& \impmed{\textbf{34.74}}
& \impsmall{\textbf{18.60}}
& \impsmall{\textbf{16.34}}
& \impmed{\textbf{12.87}}
& \impmed{\textbf{11.44}} \\
& $R_{\text{EE}} + R_{\text{fmt}}$
& 43.33 & 30.52 & 18.09 & 15.99 & 11.54 & 10.23 \\
& \hspace{3mm}+$R_{\text{span}}$
& \degsmall{42.60}
& \degsmall{29.27}
& \impsmall{18.22}
& \impsmall{16.15}
& \impsmall{11.71}
& \impsmall{10.46} \\
& \hspace{3mm}+$R_{\text{ovr}}$
& \degmed{39.02}
& \degsmall{29.39}
& \deglarge{14.56}
& \deglarge{12.92}
& \degsmall{10.52}
& \degsmall{9.54} \\
& \hspace{3mm}+$R_{\text{grd}}$
& \degsmall{41.78}
& \degsmall{29.68}
& \deglarge{14.93}
& \degmed{13.70}
& \degsmall{10.38}
& \degsmall{9.62} \\
& \hspace{3mm}+$R_{\text{cov}}$
& \degsmall{42.69}
& \degsmall{29.68}
& \impsmall{18.58}
& \impsmall{16.23}
& \impsmall{11.78}
& \impsmall{10.25} \\
\midrule

\multirow{6}{*}{M2E2}
& All
& \implarge{\textbf{67.43}}
& \implarge{\textbf{64.53}}
& \implarge{\textbf{41.21}}
& \implarge{\textbf{37.91}}
& \implarge{\textbf{33.90}}
& \implarge{\textbf{31.56}} \\
& $R_{\text{EE}} + R_{\text{fmt}}$
& 56.26 & 50.24 & 24.25 & 17.65 & 19.03 & 13.91 \\
& \hspace{3mm}+$R_{\text{span}}$
& \impmed{58.09}
& \impmed{53.50}
& \implarge{29.25}
& \implarge{23.19}
& \implarge{22.99}
& \implarge{18.54} \\
& \hspace{3mm}+$R_{\text{ovr}}$
& \implarge{62.35}
& \implarge{58.71}
& \implarge{34.45}
& \implarge{29.18}
& \implarge{27.26}
& \implarge{23.27} \\
& \hspace{3mm}+$R_{\text{grd}}$
& \implarge{60.82}
& \implarge{56.77}
& \implarge{31.27}
& \implarge{26.09}
& \implarge{24.53}
& \implarge{20.82} \\
& \hspace{3mm}+$R_{\text{cov}}$
& \impsmall{56.79}
& \impsmall{50.65}
& \impmed{26.17}
& \impmed{20.20}
& \impsmall{19.85}
& \impmed{15.46} \\
\bottomrule
\end{tabular}
\label{tab:detailed_TI_TC_AI_AC_results}
\vspace{-5mm}   
\end{table}

\paragraph{Evaluation Datasets}
To evaluate the proposed approach, we conducted experiments on 7 standard end-to-end EE datasets of different domains: WikiEvents~\cite{li2021document}, PHEE~\cite{sun2022phee}, CASIE~\cite{satyapanich2020casie}, Genia2011~\cite{kim2011overview}, Genia2013~\cite{kim2013genia}, MLEE~\cite{pyysalo2012event}, and M2E2~\cite{li2020cross}. We follow standard splits  as in \cite{huang2024textee} where we use the “split 1” data split. See Appendix~\ref{appendix:Datasets Details} and Table~\ref{tab:dataset_statistics} for more details.

\paragraph{Evaluation Metrics}
Following previous work \cite{huang2024textee,sainz2024gollie,srivastava-etal-2025-instruction}, we report F1 scores for both trigger- and argument-level identification and classification subtasks including respectively \textbf{TI}, \textbf{TC}),  \textbf{AI} and \textbf{AC}. We also report the attached version  \textbf{AI+}, \textbf{AC+}. 


We report Mean-F1 over the six subtask metrics as the main aggregate measure of end-to-end extraction quality. We additionally report full argument and trigger extraction results in Appendix~\ref{appendix:additional_results}.

\section{Results and Discussion}

\begin{figure}[ht]
    \centering
    \includegraphics[width=\linewidth]{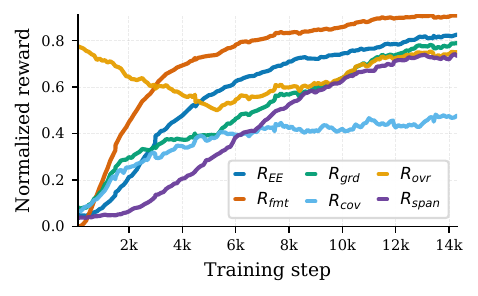}
    \caption{Training rewards dynamics. }
    \label{fig:approach_overview}
\vspace{-3mm}   
\label{fig:Training_rewards_dynamics}
\end{figure}

\subsection{Performance Analysis}
As shown in Table~\ref{tab:ee_results}, despite their strong general reasoning capabilities, frontier models such as GPT-4o (9.20), o1 (9.52), and GPT-5.4 (9.47) evaluated under few-shot prompting lag far behind task-adapted smaller models. Even the best performing GPT-5.4-mini (11.70) fails to approach the performance of the smallest GoLLIE-7B baseline (16.35), underscoring the difficulty of end-to-end event extraction and confirming  the finding in \cite{huang2024textee} that generative event extraction requires task-specific alignment beyond prompting.
Moreover, SFT provides marginal gains over zero-shot suggesting that
SFT saturates quickly and does not adequately address structured extraction in cross-domain and -schema settings. Comparing SFT-based approaches against our RL-trained model (+14.28) confirms the advantage of reinforcement learning for generative EE. Finally, \textsc{EAGER} achieves the highest average Mean-F1 of 30.79, surpassing the strongest prior RL baseline by +5.15. The gain is consistent across datasets demonstrating that the proposed approach generalizes well across diverse domains and annotation schemes. 
\paragraph{Impact of Annotation Guidelines.}
Table~\ref{tab:A.G_ablation} shows that removing annotation guidelines from \textsc{EAGER} reduces average F1 from 30.79 to 22.61  and GoLLIE-7B from 16.35 to 5.81 which is consistent with the hypothesis that annotation guidelines provide the descriptive grounding needed for schema generalization \citep{srivastava-etal-2025-instruction,
sainz2024gollie}.

\subsection{Ablation Study}
We organize our ablation evaluation around the following research questions.
\paragraph{RQ1: How can EE task-specific reward signals be effectively leveraged for RL ?}
While verifiable reward functions provide task-specific supervision across complementary aspects of event extraction, Table~\ref{tab:EAGER_ablation} shows that reward design alone is insufficient. This is reflected in the substantial performance drop from 30.79 to 22.54 average F1 when SCAE is removed. 
The same ablation applied to the $R_\text{EE} + R_\text{fmt}$ configuration (27.36 vs.\ 23.56) confirms that the benefit of SCAE is not due to the richer reward set. As shown in Figure~\ref{fig:acr}, by introducing structural diversity through varying negative schemas across sampled completions, SCAE increases advantage variance and enables more meaningful gradient updates, effectively activating the fine-grained reward signals. This suggests that without SCAE, grouped policy optimization frequently produces homogeneous outputs that prevent the model from effectively leveraging these training signals. These findings highlight that in structured extraction tasks, informative reward modeling must be coupled with optimization strategies that preserve reward diversity.

\begin{table}[t]
\centering
\scriptsize
\setlength{\tabcolsep}{1.8pt}
\renewcommand{\arraystretch}{1.18}

\caption{Ablation study on the contribution of SCAE.  Improvement/Degradations are highlighted relative to the corresponding configuration with SCAE.}
\resizebox{\columnwidth}{!}{
\begin{tabular}{@{}lcccccccc@{}}
\toprule
Setting  & Wiki & PHEE & CASIE & G11 & G13 & MLEE & M2E2 & Avg \\
\midrule

EAGER
& \textbf{21.31}
& \textbf{60.36}
& 14.66
& \textbf{26.32}
& 23.39
& \textbf{23.42}
& \textbf{46.09}
& \textbf{30.79} \\

\hspace{3mm}w/o \textbf{SCAE}
& \deglarge{13.98}
& \deglarge{50.69}
& \deglarge{8.67}
& \deglarge{15.72}
& \deglarge{15.24}
& \deglarge{12.22}
& \degmed{41.24}
& \deglarge{22.54} \\

EAGER$_{\textit{$R_{\text{EE}}$} \textit{+ $R_{\text{FMT}}$}}$
& 20.12
& 53.66
& \textbf{15.70}
& 26.11
& \textbf{24.10}
& 21.62
& 30.22
& 27.36 \\

\hspace{3mm}w/o \textbf{SCAE}
& \deglarge{15.49}
& \degsmall{52.32}
& \deglarge{9.03}
& \deglarge{16.44}
& \deglarge{16.52}
& \deglarge{11.67}
& \implarge{43.46}
& \degmed{23.56} \\

\bottomrule
\end{tabular}
}
\label{tab:EAGER_ablation}
\vspace{-3mm}
\end{table}

\paragraph{RQ2: How does each reward contribute to extraction quality?}
Starting from the baseline $R_\text{EE} + R_\text{fmt}$ (27.36 avg.\
F1), we can see in Table~\ref{tab:reward_ablation} that adding any individual specialized reward improves overall
performance, with the full combination of all six components reaching
the best performance. In contrast, Table~\ref{tab:detailed_TI_TC_AI_AC_results} shows that 
trigger metrics are relatively stable across reward configurations, as the 
$R_{\text{EE}} + R_{\text{fmt}}$ baseline already provides a reasonable foundation for trigger extraction while argument metrics are far more sensitive to reward design. Table~\ref{tab:trigger_vs_arg_summary} further reports the average gain of each additional reward  over the $R_\text{EE}+R_\text{fmt}$ baseline, separately for trigger and argument subtasks, averaged across all seven datasets.

\begin{table}[h]
	\centering
	\scriptsize
	\setlength{\tabcolsep}{5pt}
	\begin{tabular}{lcc}
		\toprule
		\textbf{Reward Added} & \textbf{Avg.\ $\Delta$ Trig.} & \textbf{Avg.\ $\Delta$ Arg.} \\
		\midrule
		$+R_\text{grd}$  & $+0.3$  & $+1.6$ \\
		$+R_\text{ovr}$  & $-0.4$  & $+3.1$ \\
		$+R_\text{cov}$  & $+0.7$  & $+0.5$ \\
		$+R_\text{span}$ & $+0.9$  & $+4.2$ \\
		\textbf{All}     & $+2.1$  & $+7.4$ \\
		\bottomrule
	\end{tabular}
	\caption{Average gain over the $R_\text{EE}+R_\text{fmt}$ baseline
		for trigger-level \textbf{Avg.\ $\Delta$ Trig.} (TI + TC) and argument-level \textbf{Avg.\ $\Delta$ Arg.} ( $\sum$ AI,AC,AI+,AC+) subtasks, averaged across seven datasets.}
	\label{tab:trigger_vs_arg_summary}
\vspace{-3mm}     
\end{table}

\noindent The results suggest that aggregate Mean-F1 gains reported in Table~\ref{tab:ee_results} are driven by argument-level improvements. Indeed, the full reward against the $R_\text{EE}+R_\text{fmt}$ configuration raises argument metrics by 7.4 pts on average versus only 2.1 pts for trigger metrics, suggesting the importance of EE tailored reward design. Additionally, no single reward dominates both subtasks simultaneously: $R_\text{cov}$ and $R_\text{span}$ are the best single-reward choices for triggers and arguments respectively, yet their combination in the full reward yields the best gains validating the proposed reward framework. Figure~\ref{fig:Training_rewards_dynamics} reveals complementary reward learning dynamics: $R_\text{fmt}$ converges rapidly while $R_\text{grd}$, $R_\text{cov}$, and $R_\text{span}$ exhibit gradual improvements throughout training.  This asymmetry
suggests that structural validity is a prerequisite condition that is
quickly satisfied, after which the model shifts its optimization toward semantic precision.

\subsection{Error Analysis}
\label{sec:error_analysis}

\begin{figure}[ht]
    \centering
    \includegraphics[width=\linewidth]{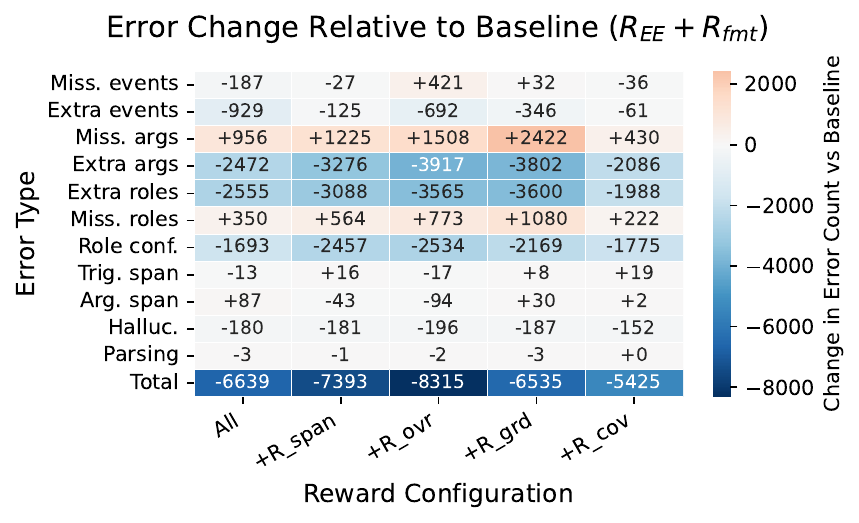}
    \caption{Error Change Relative to Baseline ($R_{EE}+R_{fmt}$). }
    \label{fig:error_delta_heatmap}
\vspace{-5mm}   
\label{fig:Training_rewards_dynamics}
\end{figure}

Figure~\ref{fig:error_delta_heatmap} reports error changes relative to the
baseline ($R_\text{EE}+R_\text{fmt}$), revealing that the dominant
error category of the baseline is over-generation. It is worth noting that the list of error categories\footnote{Error categories are defined in Table~\ref{tab:error_categories}.} is not exhaustive and does not cover all possible error types. Moreover, all reward configurations substantially reduce extra arguments ($-2{,}086$ to
$-3{,}917$), extra roles ($-1{,}988$ to $-3{,}600$), role confusion
($-1{,}693$ to $-2{,}534$), and hallucinations ($-152$ to $-196$). However, every configuration trades these gains for an increase in missing arguments ($+430$ to $+2{,}422$) and missing roles ($+222$ to
$+1{,}080$), exposing a consistent precision-recall tension across all reward designs. $R_\text{ovr}$ achieves the largest total error
reduction ($-8{,}315$) by most aggressively suppressing false positives, while $R_\text{grd}$ most severely increases missing
arguments ($+2{,}422$), reflecting over-conservative extraction. The full reward configuration (All) balances these competing pressures, yielding the second-largest total
reduction ($-6{,}639$) while keeping missing argument growth lower ($+956$) than any individual precision-focused reward alone. Span boundary and parsing errors remain marginal and stable across all configurations, confirming that low-level structural quality is not a primary bottleneck.

\section{Conclusion}

We presented EAGER, a RL framework for end-to-end event extraction that combines task-aligned verifiable rewards with SCAE. By decomposing reward supervision into complementary extraction objectives and introducing schema-contrastive grouping to mitigate reward variance collapse, our approach provides more informative optimization signals for structured extraction tasks. Experimental results across seven EE benchmark datasets demonstrate consistent improvements over strong baselines, particularly on argument-level extraction quality. Our findings further highlight the importance of combining fine-grained reward modeling with optimization strategies that preserve reward diversity in generative information extraction.

\clearpage
\section{Limitations}
Our approach relies on automatically generated annotation guidelines whose quality may vary across LLMs, domains and schemas. Investigating unified and hierarchical event schema across datasets may reduce annotation inconsistencies and improve cross-domain transfer, enabling models to better generalize across heterogeneous event definitions. 

\noindent Additionally, we  primarily assessed error categories that captures surface-level extraction anomaly and does not fully model complex phenomena such as coreference. Incorporating coreference modeling may help address cross-sentence arguments and nested event structures that remain challenging for generative EE systems. 

\noindent While the proposed framework improves event extraction performance, balancing precision and recall remains challenging. Investigating curriculum-based reinforcenment learning could better balance precision and recall during optimization.

\noindent Finally, our evaluation remains limited to  English event extraction with predefined schemas, leaving multilingual and open-schema settings for future work.

\section*{Acknowledgment}
This work was funded by the Federal Ministry of Research, Technology and Space (BMFTR) under grant number 16IW24006 (NoIDLEChatGPT) and grant number 25361 (RV-NI-2024–2029-K-IML), the Lower Saxony Ministry of Science and Culture (MWK) in the zukunft.niedersachsen program, and the Endowed Chair of AAI at University of Oldenburg. We also gratefully acknowledge support from the hessian.AI Service Center (funded by the Federal Ministry of Research, Technology and Space, BMFTR, grant no. 16IS22091) and the hessian.AI Innovation Lab (funded by the Hessian Ministry for Digital Strategy and Innovation, grant no. S-DIW04/0013/003).

\bibliography{custom}

@inproceedings{jiao2023instruct,
	title={Instruct and Extract: Instruction Tuning for On-Demand Information Extraction},
	author={Jiao, Yizhu and Zhong, Ming and Li, Sha and Zhao, Ruining and Ouyang, Siru and Ji, Heng and Han, Jiawei},
	booktitle={Proceedings of the 2023 Conference on Empirical Methods in Natural Language Processing},
	pages={10030--10051},
	year={2023}
}

@article{wang2023instructuie,
	title={Instructuie: Multi-task instruction tuning for unified information extraction},
	author={Wang, Xiao and Zhou, Weikang and Zu, Can and Xia, Han and Chen, Tianze and Zhang, Yuansen and Zheng, Rui and Ye, Junjie and Zhang, Qi and Gui, Tao and others},
	journal={arXiv preprint arXiv:2304.08085},
	year={2023}
}

@inproceedings{sainz2024gollie,
  title={Gollie: Annotation guidelines improve zero-shot information-extraction},
  author={Sainz, Oscar and Garc{\'\i}a-Ferrero, Iker and Agerri, Rodrigo and Lacalle, Oier and Rigau, German and Agirre, Eneko},
  booktitle={International Conference on Learning Representations},
  volume={2024},
  pages={47083--47107},
  year={2024}
}

@article{lu2023pivoine,
	title={Pivoine: Instruction tuning for open-world information extraction},
	author={Lu, Keming and Pan, Xiaoman and Song, Kaiqiang and Zhang, Hongming and Yu, Dong and Chen, Jianshu},
	journal={arXiv preprint arXiv:2305.14898},
	year={2023}
}

@inproceedings{huang2024textee,
	title={TextEE: Benchmark, Reevaluation, Reflections, and Future Challenges in Event Extraction},
	author={Huang, Kuan-Hao and Hsu, I-Hung and Parekh, Tanmay and Xie, Zhiyu and Zhang, Zixuan and Natarajan, Prem and Chang, Kai-Wei and Peng, Nanyun and Ji, Heng},
	booktitle={Findings of the Association for Computational Linguistics ACL 2024},
	pages={12804--12825},
	year={2024}
}

@inproceedings{chen2024large,
	title={Is a large language model a good annotator for event extraction?},
	author={Chen, Ruirui and Qin, Chengwei and Jiang, Weifeng and Choi, Dongkyu},
	booktitle={Proceedings of the AAAI conference on artificial intelligence},
	volume={38},
	number={16},
	pages={17772--17780},
	year={2024}
}

@inproceedings{li2023codeie,
	title={CodeIE: Large Code Generation Models are Better Few-Shot Information Extractors},
	author={Li, Peng and Sun, Tianxiang and Tang, Qiong and Yan, Hang and Wu, Yuanbin and Huang, Xuanjing and Qiu, Xipeng},
	booktitle={The 61st Annual Meeting Of The Association For Computational Linguistics},
	year={2023}
}

@inproceedings{li2024knowcoder,
	title={KnowCoder: Coding Structured Knowledge into LLMs for Universal Information Extraction},
	author={Li, Zixuan and Zeng, Yutao and Zuo, Yuxin and Ren, Weicheng and Liu, Wenxuan and Su, Miao and Guo, Yucan and Liu, Yantao and Lixiang, Lixiang and Hu, Zhilei and others},
	booktitle={Proceedings of the 62nd Annual Meeting of the Association for Computational Linguistics (Volume 1: Long Papers)},
	pages={8758--8779},
	year={2024}
}

@inproceedings{srivastava-etal-2025-instruction,
	title = "Instruction-Tuning {LLM}s for Event Extraction with Annotation Guidelines",
	author = "Srivastava, Saurabh  and
	Pati, Sweta  and
	Yao, Ziyu",
	editor = "Che, Wanxiang  and
	Nabende, Joyce  and
	Shutova, Ekaterina  and
	Pilehvar, Mohammad Taher",
	booktitle = "Findings of the Association for Computational Linguistics: ACL 2025",
	month = jul,
	year = "2025",
	address = "Vienna, Austria",
	publisher = "Association for Computational Linguistics",
	url = "https://aclanthology.org/2025.findings-acl.677/",
	doi = "10.18653/v1/2025.findings-acl.677",
	pages = "13055--13071",
	ISBN = "979-8-89176-256-5"
}

@article{gao2024eventrl,
	title={Eventrl: Enhancing event extraction with outcome supervision for large language models},
	author={Gao, Jun and Zhao, Huan and Wang, Wei and Yu, Changlong and Xu, Ruifeng},
	journal={arXiv preprint arXiv:2402.11430},
	year={2024}
}

@inproceedings{gui2024iepile,
	title={IEPile: Unearthing Large Scale Schema-Conditioned Information Extraction Corpus},
	author={Gui, Honghao and Yuan, Lin and Ye, Hongbin and Zhang, Ningyu and Sun, Mengshu and Liang, Lei and Chen, Huajun},
	booktitle={Proceedings of the 62nd Annual Meeting of the Association for Computational Linguistics (Volume 2: Short Papers)},
	pages={127--146},
	year={2024}
}

@inproceedings{qi2024adelie,
	title={ADELIE: Aligning Large Language Models on Information Extraction},
	author={Qi, Yunjia and Peng, Hao and Wang, Xiaozhi and Xu, Bin and Hou, Lei and Li, Juanzi},
	booktitle={Proceedings of the 2024 Conference on Empirical Methods in Natural Language Processing},
	pages={7371--7387},
	year={2024}
}

@inproceedings{wang2023code4struct,
	title={Code4Struct: Code Generation for Few-Shot Event Structure Prediction},
	author={Wang, Xingyao and Li, Sha and Ji, Heng},
	booktitle={The 61st Annual Meeting Of The Association For Computational Linguistics},
	year={2023}
}

@article{shao2024deepseekmath,
	title={Deepseekmath: Pushing the limits of mathematical reasoning in open language models, 2024},
	author={Shao, Zhihong and Wang, Peiyi and Zhu, Qihao and Xu, Runxin and Song, Junxiao and Bi, Xiao and Zhang, Haowei and Zhang, Mingchuan and Li, YK and Wu, Y and others},
	journal={URL https://arxiv. org/abs/2402.03300},
	volume={2},
	number={3},
	pages={5},
	year={2024}
}

@article{schulman2017proximal,
	title={Proximal policy optimization algorithms},
	author={Schulman, John and Wolski, Filip and Dhariwal, Prafulla and Radford, Alec and Klimov, Oleg},
	journal={arXiv preprint arXiv:1707.06347},
	year={2017}
}

@inproceedings{li2021document,
	title={Document-Level Event Argument Extraction by Conditional Generation},
	author={Li, Sha and Ji, Heng and Han, Jiawei},
	booktitle={Proceedings of the 2021 Conference of the North American Chapter of the Association for Computational Linguistics: Human Language Technologies},
	pages={894--908},
	year={2021}
}

@inproceedings{sun2022phee,
	title={PHEE: A Dataset for Pharmacovigilance Event Extraction from Text},
	author={Sun, Zhaoyue and Li, Jiazheng and Pergola, Gabriele and Wallace, Byron C and John, Bino and Greene, Nigel and Kim, Joseph and He, Yulan},
	booktitle={Proceedings of the 2022 Conference on Empirical Methods in Natural Language Processing},
	pages={5571--5587},
	year={2022}
}

@inproceedings{satyapanich2020casie,
	title={Casie: Extracting cybersecurity event information from text},
	author={Satyapanich, Taneeya and Ferraro, Francis and Finin, Tim},
	booktitle={Proceedings of the AAAI conference on artificial intelligence},
	volume={34},
	number={05},
	pages={8749--8757},
	year={2020}
}

@inproceedings{kim2013genia,
	title={The genia event extraction shared task, 2013 edition-overview},
	author={Kim, Jin-Dong and Wang, Yue and Yasunori, Yamamoto},
	booktitle={Proceedings of the BioNLP shared task 2013 workshop},
	pages={8--15},
	year={2013}
}

@article{roziere2023code,
	title={Code llama: Open foundation models for code},
	author={Roziere, Baptiste and Gehring, Jonas and Gloeckle, Fabian and Sootla, Sten and Gat, Itai and Tan, Xiaoqing Ellen and Adi, Yossi and Liu, Jingyu and Sauvestre, Romain and Remez, Tal and others},
	journal={arXiv preprint arXiv:2308.12950},
	year={2023}
}

@inproceedings{
	hu2021lora,
	title={Lo{RA}: Low-Rank Adaptation of Large Language Models},
	author={Edward J Hu and yelong shen and Phillip Wallis and Zeyuan Allen-Zhu and Yuanzhi Li and Shean Wang and Lu Wang and Weizhu Chen},
	booktitle={International Conference on Learning Representations},
	year={2022},
	url={https://openreview.net/forum?id=nZeVKeeFYf9}
}

@misc{vonwerra2022trl,
	author = {Leandro von Werra and Younes Belkada and Lewis Tunstall and Edward Beeching and Tristan Thrush and Nathan Lambert and Shengyi Huang and Kashif Rasul and Quentin Gallouédec},
	title = {TRL: Transformer Reinforcement Learning},
	year = {2020},
	publisher = {GitHub},
	journal = {GitHub repository},
	howpublished = {\url{https://github.com/huggingface/trl}}
}

@article{grattafiori2024llama,
	title={The llama 3 herd of models},
	author={Grattafiori, Aaron and Dubey, Abhimanyu and Jauhri, Abhinav and Pandey, Abhinav and Kadian, Abhishek and Al-Dahle, Ahmad and Letman, Aiesha and Mathur, Akhil and Schelten, Alan and Vaughan, Alex and others},
	journal={arXiv preprint arXiv:2407.21783},
	year={2024}
}

@inproceedings{kim2011overview,
	title={Overview of genia event task in bionlp shared task 2011},
	author={Kim, Jin-Dong and Wang, Yue and Takagi, Toshihisa and Yonezawa, Akinori},
	booktitle={Proceedings of BioNLP shared task 2011 workshop},
	pages={7--15},
	year={2011}
}

@article{pyysalo2012event,
	title={Event extraction across multiple levels of biological organization},
	author={Pyysalo, Sampo and Ohta, Tomoko and Miwa, Makoto and Cho, Han-Cheol and Tsujii, Jun'ichi and Ananiadou, Sophia},
	journal={Bioinformatics},
	volume={28},
	number={18},
	pages={i575--i581},
	year={2012},
	publisher={Oxford University Press}
}

@article{li2020cross,
	title={Cross-media structured common space for multimedia event extraction},
	author={Li, Manling and Zareian, Alireza and Zeng, Qi and Whitehead, Spencer and Lu, Di and Ji, Heng and Chang, Shih-Fu},
	journal={arXiv preprint arXiv:2005.02472},
	year={2020}
}

@inproceedings{li-etal-2023-intra,
	title = "Intra-Event and Inter-Event Dependency-Aware Graph Network for Event Argument Extraction",
	author = "Li, Hao  and
	Cao, Yanan  and
	Ren, Yubing  and
	Fang, Fang  and
	Zhang, Lanxue  and
	Li, Yingjie  and
	Wang, Shi",
	editor = "Bouamor, Houda  and
	Pino, Juan  and
	Bali, Kalika",
	booktitle = "Findings of the Association for Computational Linguistics: EMNLP 2023",
	month = dec,
	year = "2023",
	address = "Singapore",
	publisher = "Association for Computational Linguistics",
	url = "https://aclanthology.org/2023.findings-emnlp.421/",
	doi = "10.18653/v1/2023.findings-emnlp.421",
	pages = "6362--6372"
}

@inproceedings{lu-etal-2021-text2event,
    title = "{T}ext2{E}vent: Controllable Sequence-to-Structure Generation for End-to-end Event Extraction",
    author = "Lu, Yaojie  and
      Lin, Hongyu  and
      Xu, Jin  and
      Han, Xianpei  and
      Tang, Jialong  and
      Li, Annan  and
      Sun, Le  and
      Liao, Meng  and
      Chen, Shaoyi",
    editor = "Zong, Chengqing  and
      Xia, Fei  and
      Li, Wenjie  and
      Navigli, Roberto",
    booktitle = "Proceedings of the 59th Annual Meeting of the Association for Computational Linguistics and the 11th International Joint Conference on Natural Language Processing (Volume 1: Long Papers)",
    month = aug,
    year = "2021",
    address = "Online",
    publisher = "Association for Computational Linguistics",
    url = "https://aclanthology.org/2021.acl-long.217/",
    doi = "10.18653/v1/2021.acl-long.217",
    pages = "2795--2806"   
}

@inproceedings{hsu2022degree,
  title={DEGREE: A data-efficient generation-based event extraction model},
  author={Hsu, I-Hung and Huang, Kuan-Hao and Boschee, Elizabeth and Miller, Scott and Natarajan, Prem and Chang, Kai-Wei and Peng, Nanyun},
  booktitle={Proceedings of the 2022 Conference of the North American Chapter of the Association for Computational Linguistics: Human Language Technologies},
  pages={1890--1908},
  year={2022}
}

@inproceedings{ma2022prompt,
  title={Prompt for extraction? PAIE: Prompting argument interaction for event argument extraction},
  author={Ma, Yubo and Wang, Zehao and Cao, Yixin and Li, Mukai and Chen, Meiqi and Wang, Kun and Shao, Jing},
  booktitle={Proceedings of the 60th Annual Meeting of the Association for Computational Linguistics (Volume 1: Long Papers)},
  pages={6759--6774},
  year={2022}
}

@inproceedings{ren2023retrieve,
  title={Retrieve-and-sample: Document-level event argument extraction via hybrid retrieval augmentation},
  author={Ren, Yubing and Cao, Yanan and Guo, Ping and Fang, Fang and Ma, Wei and Lin, Zheng},
  booktitle={Proceedings of the 61st Annual Meeting of the Association for Computational Linguistics (Volume 1: Long Papers)},
  pages={293--306},
  year={2023}
}

@inproceedings{cai2024improving,
  title={Improving event definition following for zero-shot event detection},
  author={Cai, Zefan and Kung, Po-Nien and Suvarna, Ashima and Ma, Mingyu and Bansal, Hritik and Chang, Baobao and Brantingham, P Jeffrey and Wang, Wei and Peng, Nanyun},
  booktitle={Proceedings of the 62nd Annual Meeting of the Association for Computational Linguistics (Volume 1: Long Papers)},
  pages={2842--2863},
  year={2024}
}

@inproceedings{hong2024towards,
  title={Towards better question generation in QA-based event extraction},
  author={Hong, Zijin and Liu, Jian},
  booktitle={Findings of the Association for Computational Linguistics: ACL 2024},
  pages={9025--9038},
  year={2024}
}

@inproceedings{ma2024star,
  title={STAR: boosting low-resource information extraction by structure-to-text data generation with large language models},
  author={Ma, Mingyu Derek and Wang, Xiaoxuan and Kung, Po-Nien and Brantingham, P Jeffrey and Peng, Nanyun and Wang, Wei},
  booktitle={Proceedings of the AAAI conference on artificial intelligence},
  volume={38},
  number={17},
  pages={18751--18759},
  year={2024}
}

@article{huan2025does,
  title={Does math reasoning improve general llm capabilities? understanding transferability of llm reasoning},
  author={Huan, Maggie and Li, Yuetai and Zheng, Tuney and Xu, Xiaoyu and Kim, Seungone and Du, Minxin and Poovendran, Radha and Neubig, Graham and Yue, Xiang},
  journal={arXiv preprint arXiv:2507.00432},
  year={2025}
}

@article{chu2025sft,
  title={Sft memorizes, rl generalizes: A comparative study of foundation model post-training},
  author={Chu, Tianzhe and Zhai, Yuexiang and Yang, Jihan and Tong, Shengbang and Xie, Saining and Schuurmans, Dale and Le, Quoc V and Levine, Sergey and Ma, Yi},
  journal={arXiv preprint arXiv:2501.17161},
  year={2025}
}

@article{gao2023exploring,
  title={Exploring the feasibility of chatgpt for event extraction},
  author={Gao, Jun and Zhao, Huan and Yu, Changlong and Xu, Ruifeng},
  journal={arXiv preprint arXiv:2303.03836},
  year={2023}
}

@article{hurst2024gpt,
  title={Gpt-4o system card},
  author={Hurst, Aaron and Lerer, Adam and Goucher, Adam P and Perelman, Adam and Ramesh, Aditya and Clark, Aidan and Ostrow, AJ and Welihinda, Akila and Hayes, Alan and Radford, Alec and others},
  journal={arXiv preprint arXiv:2410.21276},
  year={2024}
}

@article{yu2026dapo,
  title={Dapo: An open-source llm reinforcement learning system at scale},
  author={Yu, Qiying and Zhang, Zheng and Zhu, Ruofei and Yuan, Yufeng and Zuo, Xiaochen and Yue, Yu and Dai, Weinan and Fan, Tiantian and Liu, Gaohong and Liu, Lingjun and others},
  journal={Advances in Neural Information Processing Systems},
  volume={38},
  pages={113222--113244},
  year={2026}
}

@article{ouyang2022training,
  title={Training language models to follow instructions with human feedback},
  author={Ouyang, Long and Wu, Jeffrey and Jiang, Xu and Almeida, Diogo and Wainwright, Carroll and Mishkin, Pamela and Zhang, Chong and Agarwal, Sandhini and Slama, Katarina and Ray, Alex and others},
  journal={Advances in neural information processing systems},
  volume={35},
  pages={27730--27744},
  year={2022}
}

@article{guo2025deepseek,
  title={Deepseek-r1: Incentivizing reasoning capability in llms via reinforcement learning},
  author={Guo, Daya and Yang, Dejian and Zhang, Haowei and Song, Junxiao and Wang, Peiyi and Zhu, Qihao and Xu, Runxin and Zhang, Ruoyu and Ma, Shirong and Bi, Xiao and others},
  journal={arXiv preprint arXiv:2501.12948},
  year={2025}
}

@article{singh2025openai,
  title={Openai gpt-5 system card},
  author={Singh, Aaditya and Fry, Adam and Perelman, Adam and Tart, Adam and Ganesh, Adi and El-Kishky, Ahmed and McLaughlin, Aidan and Low, Aiden and Ostrow, AJ and Ananthram, Akhila and others},
  journal={arXiv preprint arXiv:2601.03267},
  year={2025}
}

@article{jaech2024openai,
  title={Openai o1 system card},
  author={Jaech, Aaron and Kalai, Adam and Lerer, Adam and Richardson, Adam and El-Kishky, Ahmed and Low, Aiden and Helyar, Alec and Madry, Aleksander and Beutel, Alex and Carney, Alex and others},
  journal={arXiv preprint arXiv:2412.16720},
  year={2024}
}

@inproceedings{he2026avspo,
  title     = {Advantage Collapse in Group Relative Policy Optimization: Diagnosis and Mitigation},
  author    = {He, Xixiang and Sun, Qiyao and Cheng, Ao and Li, Xingming and Ji, Xuanyu and Lu, Hailun and Huang, Runke and Hu, Qingyong},
  booktitle = {International Conference on Machine Learning},
  year      = {2026}
}

@article{zhang2025scaf,
  title={Scaf-grpo: Scaffolded group relative policy optimization for enhancing LLM reasoning},
  author={Zhang, Xichen and Wu, Sitong and Zhu, Yinghao and Tan, Haoru and Yu, Shaozuo and He, Ziyi and Jia, Jiaya},
  journal={arXiv preprint arXiv:2510.19807},
  year={2025}
}

@article{liu2026gdpo,
  title={Gdpo: Group reward-decoupled normalization policy optimization for multi-reward rl optimization},
  author={Liu, Shih-Yang and Dong, Xin and Lu, Ximing and Diao, Shizhe and Belcak, Peter and Liu, Mingjie and Chen, Min-Hung and Yin, Hongxu and Wang, Yu-Chiang Frank and Cheng, Kwang-Ting and others},
  journal={arXiv preprint arXiv:2601.05242},
  year={2026}
}

@inproceedings{adjali2026aligning,
  title={Aligning Instruction-Tuned LLMs for Event Extraction with Multi-objective Reinforcement Learning},
  author={Adjali, Omar and Liang, Siting and Bhatti, Omair Shahzad and Sonntag, Daniel},
  booktitle={European Conference on Information Retrieval},
  pages={586--595},
  year={2026}
}


\appendix

\section{Code-based Representation}
\label{appendix:Code-based Representation}

\begin{figure}[b]
\begin{lstlisting}[style=pythonstyle]
@dataclass
class Conflict_Attack:
    """Conflict:Attack event"""
    mention: str
    Attacker: List[str]
    Instrument: List[str]
    Place: List[str]
    Target: List[str]
\end{lstlisting}
\vspace{-3mm}
\caption{Example of an event schema as python class}
\label{fig:event_python_class}
\end{figure}
We formulate  EE as  a code generation problem  where both the input  and output are formatted using Python code similar to Following \cite{sainz2024gollie,srivastava-etal-2025-instruction}. Indeed, IE tasks  benefit on the first hand from the strong code understanding capabilities of large language models since code data is widely included in their pre-training corpora \cite{wang2023code4struct,li2023codeie}. On the other hand, code-based structure representation provides a unified and human-readable framework for information extraction tasks, while mitigating ambiguities often encountered in natural language instructions. The code format ensures that outputs are syntactically well-formed facilitating output parsing  \cite{sainz2024gollie,srivastava-etal-2025-instruction}. In particular, event schemes are expressed  as Python classes (\texttt{@dataclass} type definitions) and the  extracted output events as instances of these classes. 

\section{Guidelines Annotation Generation} 
\label{appendix:Guidelines Annotation Augmentation}
Using LLaMA-3.1-8B-Instruct model   \cite{grattafiori2024llama}, we adopt the \textit{Guideline-PN} (Positive + Negative) generation protocol of~\cite{srivastava-etal-2025-instruction}, wherein the LLM is conditioned on a contrastive set of examples to produce guidelines for each event type $e \in \mathcal{E}$. Specifically, the prompt consists of positive examples: 10 annotated instances of event type $e$ paired with their source texts and negative examples: 15 texts containing other event types but no instance of $e$. This contrastive design encourages the model to identify definitional boundaries and discriminative features that distinguish $e$ from related event types. The generation prompt instructs the LLM to: (1) enumerate all unique argument roles for $e$; (2) provide a precise definition of the event type; and (3) characterize each argument role, emphasizing its semantic function, representative mentions, and potential edge cases. The resulting guidelines are incorporated into the Python class schema via docstrings and inline comments, yielding an augmented schema $E_e^{\text{G}}$ that integrates both structural and semantic information. See Figures~\ref{fig:WikiEvents_event_schema_exp},~\ref{fig:PHEE_event_schema_exp},~\ref{fig:CASIE_event_schema_exp},~\ref{fig:Genia2011_event_schema_exp},~\ref{fig:Genia2013_event_schema_exp},~\ref{fig:MLEE_event_schema_exp} and ~\ref{fig:M2E2_event_schema_exp} for examples of  event scheme augmented with annotation guidelines for each dataset.

\subsection{Additional Annotation Guidelines Analysis}
Tables \ref{tab:gollie_wo_ag} reports the the effect of AG on the GoLLIE backbone model. We can see that augmenting event schemes with annotation guidelines improves event extraction performance. 
\begin{table}[t]
\centering
\small
\resizebox{\columnwidth}{!}{%
\begin{tabular}{lrrrrrrrr}
\toprule
Model & WikiEvents & PHEE & CASIE & GENIA11 & GENIA13 & MLEE & M2E2 & Avg. \\
\midrule
GoLLIE-7B w/o AG  & 9.14 & 7.67 & 3.13 & 1.64 & 1.17 & 2.41 & 15.54 & 5.81 \\
GoLLIE-13B w/o AG & 10.07 & 7.78 & 4.00 & 2.16 & 0.99 & 2.88 & 22.31 & 7.17 \\
GoLLIE-34B w/o AG & 10.46 & 9.71 & 4.25 & 5.26 & 5.00 & 4.55 & 29.37 & 9.80 \\
\bottomrule
\end{tabular}%
}
\caption{Performance comparison of GoLLIE models without AG.}
\label{tab:gollie_wo_ag}
\end{table}


\section{Supervised Fine-Tuning} 
\label{appendix:prompt_sft}
To effectively train medium-scale LLMs while preserving their foundational capabilities, we carry out instruction-based supervised fine-tuning  to adapt the model to event extraction using code-style schema representations. This enables the model to follow natural-language task instructions while generating syntactically valid and semantically grounded event representations in Python code.
The LLM is trained using supervised fine-tuning on a set of annotated prompt/response pairs $\{(P_i, Y_i)\}_{i=1}^N$, where $Y_i$ denotes the target structured event instance. The fine-tuning objective follows a standard autoregressive likelihood formulation:
\[
\mathcal{L}( \theta) = - \sum_{i} \sum_{j} \log p_{\theta}(Y_{i,j} \mid P_i, Y_{i,<j}),
\]
where $Y_{i,<j}$ denotes previously generated tokens in the output sequence.

\section{Inference}
At inference time, given an input text $X$, all $n$ event schemas in $\mathcal{E} = \{E_i\}_{i=1}^{n}$ are provided jointly in the prompt, enabling the model to perform end-to-end event extraction and produce a Python list of instantiated event objects in a single forward pass.

\section{DAPO Background}
\noindent DAPO builds upon GRPO and addresses its key limitations i.e., entropy collapse, reward noise, and sequence-level length bias through four
modifications: (1) Clip-Higher, which uses asymmetric clipping bounds
$[\epsilon_{\mathrm{low}}, \epsilon_{\mathrm{high}}]$ with
$\epsilon_{\mathrm{high}} > \epsilon_{\mathrm{low}}$ to prevent entropy collapse;
(2) Token-Level Loss, which normalizes the policy gradient over all active
tokens in the batch rather than per sequence, eliminating length bias;
(3) Dynamic Sampling, which filters out zero-variance groups (all-correct
or all-incorrect) and resamples until every batch contains informative gradient
signal; and (4) KL Removal, which sets $\beta{=}0$. Unlike standard RLHF~\cite{ouyang2022training}, where KL regularization prevents the policy 
from deviating too far from a supervised baseline, EE requires
the model to diverge from its initial distribution to acquire precise schema-grounded output patterns. 

\section{Verifiable Rewards Formulation}
\label{appendix:Verifiable Rewards Formulation}
We propose a decomposed, task-aligned reward formulation for generative EE. Rather than relying solely on aggregate extraction F1, we organize the reward signal around three complementary objectives: \textit{validity}, requiring outputs to be parseable and schema-compatible; \textit{extraction accuracy}, requiring outputs to match gold event annotations; and \textit{generation behavior}, requiring outputs to remain grounded in the source text, avoid spurious predictions, and maintain adequate event coverage.

\subsection{Validity Reward}
As a prerequisite to semantic evaluation, we reward outputs that are syntactically well-formed and successfully parse into the code-based representation introduced in Section~\ref{sec:code_repr}:
\begin{equation}
  R_{\text{fmt}}(o_i) =
  \begin{cases}
    1, & \text{if } o_i \text{ is syntactically parseable,}\\
    0, & \text{otherwise.}
  \end{cases}
  \label{eq:reward-format}
\end{equation}

\subsection{Extraction Accuracy Reward}
We define the extraction reward as the sum of F1 scores across the six EE evaluation metrics spanning the four canonical subtasks TI, TC, AI, AC ‚and their trigger-aware variants AI$^{+}$ and AC$^{+}$:
\begin{equation}
  R_{\text{EE}} =
  \sum_{k \,\in\,
  \{\text{TI, TC, AI, AC, AI}^{+}\!,\,\text{AC}^{+}\}}
  F_1^k.
  \label{eq:reward-ee}
\end{equation}
This reward captures holistic extraction quality across all subtasks but remains coarse with respect to specific generation issues, motivating the supplementary signals below.

\subsection{Groundedness Reward}
To penalize hallucinated triggers and argument spans unsupported by the source text, we introduce a groundedness reward that measures whether predicted mentions appear verbatim in the input $X$. Let $\{e_j\}$ denote the set of predicted events, $\mathcal{R}(e_j)$ the set of argument roles for event $e_j$, and $T_i$ the total number of predicted arguments across all events in $o_i$. We compute separate support scores for triggers and arguments:
\begin{align}
  s_{\text{m}} &=
  \frac{1}{|\{e_j\}|}
  \sum_{e_j} \mathbb{I}\bigl[\mathrm{contains}(X,\, e_j.\mathrm{trigger})\bigr],
  \label{eq:sm}\\
  s_{\text{arg}} &=
  \frac{1}{T_i}
  \sum_{e_j}\sum_{r \in \mathcal{R}(e_j)}\sum_{v \in e_j[r]}
  \mathbb{I}\bigl[\mathrm{contains}(X,\, v)\bigr],
  \label{eq:sarg}
\end{align}
with $s_{\text{m}} = 0$ if no events are predicted and $s_{\text{arg}} = 0$ if $T_i = 0$. The groundedness reward is their average:
\begin{equation}
  R_{\text{grd}} = \tfrac{1}{2}\bigl(s_{\text{m}} + s_{\text{arg}}\bigr).
  \label{eq:reward-ground}
\end{equation}
This signal penalizes fabricated spans while remaining agnostic to event type correctness. 

\subsection{Over-generation Reward}
Since extraction F1 does not explicitly penalize spurious predictions beyond the gold 
annotation, we introduce a complementary over-generation penalty. Let $E_i^{+}$ and 
$A_i^{+}$ denote the number of predicted events and arguments in $o_i$ that exceed 
the gold counts in $Y^\ast$. We define:

\begin{equation}
  R_{\mathrm{ovr}} = \max\!\left(0,\; 1 - \frac{E^{+}_{i} + A^{+}_{i}}{n^{*} + a^{*}}\right)
  \label{eq:reward-overgen}
\end{equation}
where $E^{+}_{i}$ and $A^{+}_{i}$ are the predicted events and arguments exceeding the gold counts, and $n^{*}$, $a^{*}$ are the gold event and argument counts.
This reward equals 1 when no excess predictions are made and decreases linearly as spurious spans accumulate, saturating at 0.

\subsection{Coverage Reward}
To counterbalance $R_{\text{ovr}}$ and prevent the model from adopting an overly conservative decoding strategy, we reward proportional event and argument coverage relative to the gold annotation. Let $n^\ast = |Y^\ast|$ and $n^g = |\{e \in o_i : e.\text{type} \in \mathcal{S}\}|$ denote the gold and schema-valid predicted event counts, respectively, and let $a^\ast$ and $a^g$ be the corresponding argument counts. We define:
\begin{equation}
  R_{\text{cov}} =
  \frac{1}{2}\!\left(
    \min\!\left(1,\frac{n^g}{n^\ast}\right) +
    \min\!\left(1,\frac{a^g}{a^\ast}\right)
  \right),
  \label{eq:reward-cov}
\end{equation}
where $\mathcal{S}$ is the event schema registry. The $\min(\cdot,1)$ clamping ensures that over-prediction does not inflate the coverage score, maintaining complementarity with $R_{\text{ovr}}$.

\subsection{Span Precision Reward}
\label{sec:span_reward}
It addresses the case where the model correctly identifies an argument's semantic 
content but extracts a superset of the minimal gold span. Standard exact-match rewards 
treat such predictions as fully incorrect, providing no useful gradient signal for 
boundary refinement. We therefore introduce a complementary Span Precision Reward 
$R_{\mathrm{span}}$ that softly penalizes overpredicted spans while rewarding 
near-correct predictions. For each predicted argument $\hat{a}$, we retrieve its 
best-matching gold span $a^* = \argmax_{a \in \mathcal{A}_r}\,\mathcal{J}(\hat{a}, a)$ 
via token Jaccard similarity, and compute a per-argument score:
\begin{multline}
s(\hat{a}, a^*) =
\begin{cases}
\max\!\left(
0,\;
\mathcal{J}(\hat{a}, a^*) \right. \\
\left.
-
\dfrac{|\hat{a}|-|a^*|}{\max(1,|a^*|)}
\right)
\quad \text{if } \hat{a} \supset a^*, \\[4pt]
\mathcal{J}(\hat{a}, a^*)
\quad \text{otherwise.}
\end{cases}
\end{multline}
The reward is the mean score over all aligned predicted arguments, and is gated to 
zero when no overprediction occurs, decoupling it from the outcome reward in 
well-behaved cases.

\section{Implementation Details}
\label{appendix_training_details}
We conducted experiments using the GoLLIE-7B model~\cite{sainz2024gollie}, a fine-tuned version of Code-LLaMA~\cite{roziere2023code}, which provides a strong backbone for code-based information extraction. For efficient optimization, we employed parameter-efficient fine-tuning via QLoRA~\cite{hu2021lora}, applying low-rank adapters to the attention projection layers with LoRA rank 8, scaling factor $\alpha=16$, learning rate $1\times10^{-6}$, and batch size 1 for SFT. For RL optimization, we used the TRL library~\cite{vonwerra2022trl} with learning rate $1\times10^{-6}$, batch size 4, and 4 sampled completions per input using nucleus sampling ($p=0.9$, $\tau=0.6$). DAPO Clipping bounds are set to $\epsilon_{\text{low}}=0.2$ and $\epsilon_{\text{high}}=0.28$ following \citet{yu2026dapo}, and the KL penalty coefficient is set to $\beta=0$. Models were trained for up to 10 epochs on 4 NVIDIA H100 GPUs for RL optimization, with early stopping triggered after 3 consecutive non-improving validation steps. The best-performing validation checkpoint was selected for testing. At inference time, we used greedy decoding. Our code will be released at:\url{https://github.com/OA256864/EE_RL}.  

\section{Training Protocol}
\label{sec:training_protocol}
We train the GoLLIE-7B LLM on a training set constructed by concatenating the training sets of the seven datasets used in our experiments and then we shuffled the result set to have cross-domain and -schema training batches. We conducted DAPO \cite{yu2026dapo}  with our proposed SCAE and our reward formulation without resorting to an SFT checkpoint for warm-starting as  SFT provided only marginal improvements on the validation sets.
\paragraph{Impact of number of negative schemas K} 
Our preliminary experiments indicated that increasing the number of negative schemas K sampled using SCAE  consistently improved performance. Consequently, we set K=10, the largest value permitted by both our GPU memory constraints and the 16k-token maximum prompt length budget.

\paragraph{Impact of group size G}
Similarly, preliminary experiments showed that increasing the group size from $G=2$ to $G=4$ sampled completions per input improved performance while also accelerating convergence. These results suggest that scaling to $G=8$ could yield further performance gains given a very large computational budget.

\begin{figure*}[t]
\begin{lstlisting}[style=pythonstyle]
'''This is an event extraction task where the goal is to extract structured events from the text. A structured event contains an event trigger word, an event type, the arguments participating in the event, and their roles in the event. For each different event type, please output the extracted information from the text into python-style dictionaries where the first key will be 'mention' with the value of the event trigger. Next, please output the arguments and their roles follow ing the same format. The event type definitions and their argument roles are defined next.'''

@dataclass
class Cognitive_Inspection_SensoryObserve:
    """The event is triggered by the act of observing or inspecting something with one's senses, such as sight, sound, touch, taste, or smell. It involves the use of an instrument, such as a camera, microscope, or other device, to gather information about an entity, which can be a person, place, or object. The event is typically performed by an observer, who may be an individual or a group, and takes place at a specific location. Unlike other events, such as Justice_Convict_Unspecified or Transaction_ExchangeBuySell_Unspecified,
    Cognitive_Inspection_SensoryObserve does not involve a legal or financial transaction. Triggers such as 'searched', 'found', 'looked', or 'reviewing' are indicative of Cognitive_Inspection_SensoryObserve, not other event types."""

    mention: str
    # The mention argument represents the verb or action that triggers the
    # Cognitive_Inspection_SensoryObserve event. Examples are 'searched',
    # 'found', 'looked', 'reviewing', or 'observed'.

    Instrument: List[str]
    # The Instrument argument represents the tool or device used to gather
    # information about the observed entity. Examples are 'camera',
    # 'microscope', 'laptop computer', or 'robot'.

    ObservedEntity: List[str]
    # The ObservedEntity argument represents the person, place, or object being
    # observed. Examples are 'Martin Farnsworth', 'bombs', 'laptop computer',
    # or 'backpack'.

    Observer: List[str]
    # The Observer argument represents the person or group performing the
    # observation. Examples are 'agents', 'investigators', 'police',
    # or 'workers'.

    Place: List[str]
    # The Place argument represents the location where the observation takes
    # place. Examples are 'landfill', 'station', 'Boston', or 'pool'.

@dataclass...
# ...+ list of remaining GT Event schemes + Negative Event schemes  ...

# This is the text to analyze                                                                                                                                      
text = 'Boston Bomb Suspect Sent to Federal Medical Detention Boston Marathon bombing suspect Dzhokhar Tsarnaev has been moved to a prison medical facility as authorities continue to search for answers about the attack . The U . S . Marshals Service said Friday that Tsarnaev was moved to the Federal Medical Center Devens , a Bureau of Prisons facility in the northeastern state of Massachusetts . He was transferred there from a Boston hospital where he had been receiving treatment for injuries sustained during his capture last week . Federal Medical Center Devens , Ayer , Massachusetts A spokesman did not give details about the condition of the 19 - year - old , who officials say is recovering from a neck wound...'

# The list called result contains the event instances according to the guidelines above:

result =
\end{lstlisting}
\caption{Training Input Prompt Example}

\label{fig:example_input_prompt}
\end{figure*}

\section{Pseudo Code of DAPO with SCAE}
To provide an overview of the training process, we present the pseudo code of DAPO with SCAE in Algorithm~\ref{alg:scae_alg}.

\begin{algorithm}[t]
\caption{DAPO with SCAE}
\begin{algorithmic}[1]
\Require Initial policy $\pi_{\theta}$; training set $\mathcal{D}$; 
         event schema pool $\mathcal{E}$; group size $G$; 
         negative schema count $K$; 
         clipping bounds $\epsilon_{\mathrm{low}}, \epsilon_{\mathrm{high}}$
\For{step $= 1, \ldots, n$}
    \State Sample a batch $\mathcal{D}_b$ from $\mathcal{D}$
    \State Update old policy $\pi_{\theta_{\mathrm{old}}} \leftarrow \pi_{\theta}$
    \For{each instance $(X, e, Y) \in \mathcal{D}_b$}
        \State Let $\mathcal{N} = \mathcal{E} \setminus \{e\}$ be the negative schema pool
        \For{$i = 1, \ldots, G$}
            \State Sample a distinct negative schema subset 
                   $\mathcal{S}_i \overset{\text{i.i.d.}}{\sim} \binom{\mathcal{N}}{K}$
            \State Construct contrastive prompt 
                   $P_i = I \oplus E_e^{\mathrm{G}} \oplus \mathcal{S}_i \oplus X$
            \State Sample completion 
                   $\hat{o}_i \sim \pi_{\theta_{\mathrm{old}}}(\cdot \mid P_i)$
            \State Compute reward $R_i = R(\hat{o}_i, Y)$
        \EndFor
        \State Compute group statistics $\mu_R = \mathrm{mean}(\{R_i\}_{i=1}^{G})$,\; 
               $\sigma_R = \mathrm{std}(\{R_i\}_{i=1}^{G})$
        \If{$\sigma_R = 0$} 
            \State \textbf{skip} group (Dynamic Sampling)
        \EndIf
        \State Compute per-token advantages 
               $\hat{A}_t^i = ({R_i - \mu_R})/{\sigma_R}$
    \EndFor
    \State Update $\pi_{\theta}$ by minimizing $\mathcal{L}_{\mathrm{DAPO}}(\theta)$ 
           (Eq.~\ref{eq:dapo-objective}) with advantages $\{\hat{A}_t^i\}$
\EndFor
\State \Return $\pi_{\theta}$
\end{algorithmic}
\label{alg:scae_alg}
\end{algorithm}

\section{Training Dynamics Analysis}

\begin{figure}[H]
    \centering
    \includegraphics[width=\linewidth]{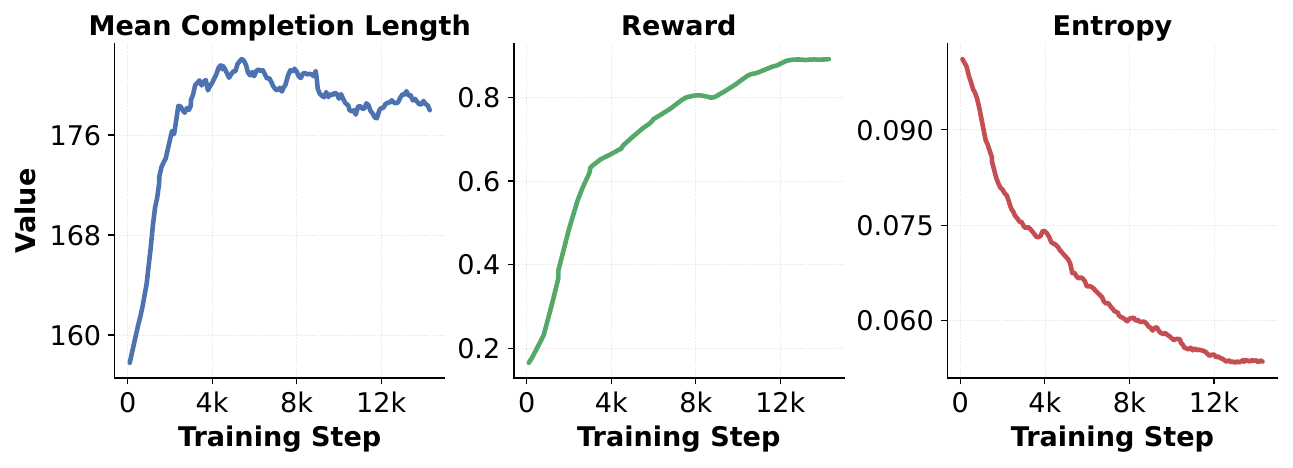}
    \caption{Analysis of the main metrics for monitoring our RL training dynamics 
    including the mean completion length, the mean reward score, and generation entropy.}
    \label{fig:training_dynamics_analysis}
\end{figure}

Reinforcement learning over structured prediction tasks such as EE introduces 
additional complexity beyond standard reasoning benchmarks, as the reward signal 
must simultaneously supervise trigger identification, event classification, and 
schema-grounded argument extraction. Given this interdependence, monitoring key 
intermediate metrics throughout training is essential for diagnosing undesired behavior 
and validating that each reward component contributes as intended. 
Figure~\ref{fig:training_dynamics_analysis} reports the three principal indicators 
we track across training in addition to the EE validation performance.

\subsection{Mean Completion Length Analysis}
The mean completion length increases steadily over the course of training, reflecting the model's growing tendency to produce more 
complete structured outputs as training progresses. In the context of code-based EE, 
this growth is consistent with the model learning to instantiate a larger and more 
complete set of event objects and argument slots, rather than defaulting to 
under-populated outputs. We observe no prolonged stagnation or decline in length, 
suggesting that our training signal discourages overly conservative decoding throughout training.

\subsection{Mean Reward Analysis}
The mean reward increases monotonically   with a smooth trajectory and no sign of instability or collapse. This stability indicates that the 
decomposed reward formulation $R_{\text{full}}$ provides a reliable and consistent 
training signal, allowing the model to robustly fit the distribution of the training set. Consistent with findings reported in prior DAPO work~\cite{yu2026dapo}, we 
observe that the final reward on the training set correlates imperfectly with 
held-out extraction performance, underscoring the importance of complementing reward monitoring with validation-set evaluation to detect potential overfitting.

\subsection{Generation Entropy Analysis}
The generation entropy exhibits a slow but consistent downward trend throughout 
training until it stabilizes then it starts increasing slightly. As shown in~\cite{yu2026dapo}, too high entropy indicates over-exploration of the model, while too low entropy leads to a loss of exploration capability suggesting that the model’s entropy needs to be maintained within an appropriate range. In our setting, an exploratory behavior  is particularly important, as the model must explore diverse span-selection and role-assignment strategies before converging on schema-grounded outputs. The absence of any entropy spike or erratic fluctuation further confirms that the removal of KL regularization ($\beta{=}0$) does not destabilize training in our setting.

\section{Training Computational Cost}
Table~\ref{tab:compute_cost_train} reports the computational training cost.
\begin{table}[t]
\centering
\small
\renewcommand{\arraystretch}{0.8}
\resizebox{\columnwidth}{!}{%
\begin{tabular}{llcccc}
\toprule
\textbf{Method} & \textbf{Model} & \textbf{GPUs} & \textbf{Batch Size} & \textbf{Epochs} & \textbf{Wall Time (h)} \\
\midrule
EAGER & GoLLIE-7B & 4$\times$H100 & 4 & 1 & $\sim$4 \\
\bottomrule
\end{tabular}%
}
\caption{Computational cost of EAGER training on 4 NVIDIA H100 GPUs. Wall time is reported for the full train dataset, which includes the seven dataset training sets.}
\label{tab:compute_cost_train}
\end{table}

\section{Datasets Details}
\label{appendix:Datasets Details}
We evaluate EAGER on seven benchmark datasets spanning diverse domains, annotation schemes, and event ontologies.

\paragraph{WikiEvents}~\cite{li2021document} is a large-scale news-domain benchmark containing richly annotated real-world event mentions with complex argument structures.

\paragraph{PHEE}~\cite{sun2022phee} focuses on the biomedical domain, specifically pharmacovigilance event extraction from medical case reports, requiring fine-grained reasoning over domain-specific terminology.

\paragraph{CASIE}~\cite{satyapanich2020casie} is a cybersecurity event extraction benchmark centered on security incident reports, featuring highly specialized event schemas and technical vocabulary.

\paragraph{GENIA2011 and GENIA2013}~\cite{kim2011overview,pyysalo2012event} are biomedical event extraction datasets derived from PubMed abstracts, involving nested event structures and biologically grounded argument semantics.

\paragraph{MLEE}~\cite{pyysalo2012event} extends biomedical event extraction to molecular-level event understanding with more diverse biological interaction types.

\paragraph{M2E2}~\cite{li2020cross} is a multimodal event extraction benchmark originally designed for joint text-image event understanding; following prior text-only work, we evaluate exclusively on the textual component.

\noindent These datasets collectively cover news, biomedical, and cybersecurity domains, providing a comprehensive testbed for evaluating robustness across heterogeneous event schemas and extraction difficulty levels.

\begin{table}[t]
\centering
\small
\setlength{\tabcolsep}{5pt}
\renewcommand{\arraystretch}{1.15}

\resizebox{\linewidth}{!}{%
\begin{tabular}{lrrrrrl}
\toprule
\textbf{Dataset} &
\textbf{\#Docs} &
\textbf{\#Instances} &
\textbf{\#Event Types} &
\textbf{\#Events} &
\textbf{\#Arguments} &
\textbf{Domain} \\
\midrule

WikiEvents & 245  & 565  & 50 & 598   & 5,501  & Wikipedia \\
CASIE      & 999  & 1,375 & 5  & 8,469 & 22,575 & Cybersecurity \\
PHEE       & 4,827 & 4,827 & 2  & 5,019 & 25,760 & Pharmacovigilance \\
GENIA2011  & 960  & 960  & 9  & 13,537 & 11,865 & Biomedical \\
GENIA2013  & 20   & 664  & 13 & 6,001 & 5,660  & Biomedical \\
MLEE       & 262  & 286  & 29 & 6,575 & 5,958  & Biomedical \\
M2E2       & 6,013 & 6,013 & 8  & 1,105 & 1,659  & News \\
\bottomrule

\end{tabular}%
}

\caption{Statistics of the end-to-end event extraction datasets used in our experiments.}
\label{tab:dataset_statistics}

\end{table}

\section{Additional Results and Analysis}
\label{appendix:additional_results}
Table~\ref{tab:full_results_WikiEvents_PHEE_CASIE}, Table~\ref{tab:full_results_genia11_13} and Table~\ref{tab:full_results_mlee_m2e2} show the detailed results of EE.

\subsection{Trigger and Argument Performance Analysis}
We decompose event extraction quality into trigger-level (TI, TC) and
argument-level (AI, AC, AI+, AC+) subtasks using the detailed results and isolating how each reward contributes to the global EE quality.

\subsubsection{Event Trigger Performance (TI,TC)}
The $R_\text{EE}+R_\text{fmt}$ baseline achieves competitive TI and TC
scores across most datasets indicating that schema-grounded
instruction tuning already provides a reasonable foundation for trigger
localisation and type assignment. $R_\text{cov}$ yields the most consistent TI gains, by penalising missed gold events and directly improving trigger recall. $R_\text{span}$ contributes primarily to TC by refining span boundaries and stabilising type-assignment confidence. $R_\text{ovr}$ introduces a precision-recall trade-off, reducing spurious triggers but degrading TI on datasets where the baseline
already under-generates. Combining all rewards yields the strongest TI and TC on most datasets, with the largest gains on M2E2 (TI: $+11.17$, TC: $+14.29$ over baseline) and MLEE (TI: $+3.22$, TC: $+4.22$), where multi-event documents benefit most from joint coverage and precision supervision.

\subsubsection{Argument Performance (AI, AC, AI+, AC+)}
Argument metrics are significantly lower than trigger metrics, and the
anchored variants AI+ and AC+ collapse further exposing systematic
trigger-argument misalignment that is not apparent using mean-F1.
$R_\text{span}$ is the single most impactful reward for argument
tasks, delivering the largest per-dataset gains across all four
subtasks (PHEE: AC $+14.15$, AC+ $+9.64$; M2E2: AC $+5.54$,
AC+ $+4.63$), as boundary-level supervision directly addresses the
span imprecision that drives argument scoring errors. $R_\text{ovr}$
provides strong corrections in over-extraction behaviors (M2E2: AI+
$+8.23$, AC+ $+9.36$) but degrades recall in domains
such as CASIE and MLEE. $R_\text{grd}$ improves AI and AC where
hallucinated spans are prevalent (WikiEvents, PHEE) but shows limited
benefit for AI+/AC+, indicating that verbatim-span enforcement alone
does not resolve trigger-argument misalignment.
When resorting to the full reward set produces the highest AI+/AC+ scores on the majority of datasets, with the most pronounced gains on M2E2 (AI+:
$+14.87$, AC+: $+17.65$ over baseline) and PHEE (AC+: $+11.02$).
These improvements suggest that the combined reward induces trigger-argument alignment beyond what individual objectives achieve in isolation.

\subsection{Additional Error Analysis}
\begin{figure*}[ht]
    \centering
    \includegraphics[width=\textwidth]{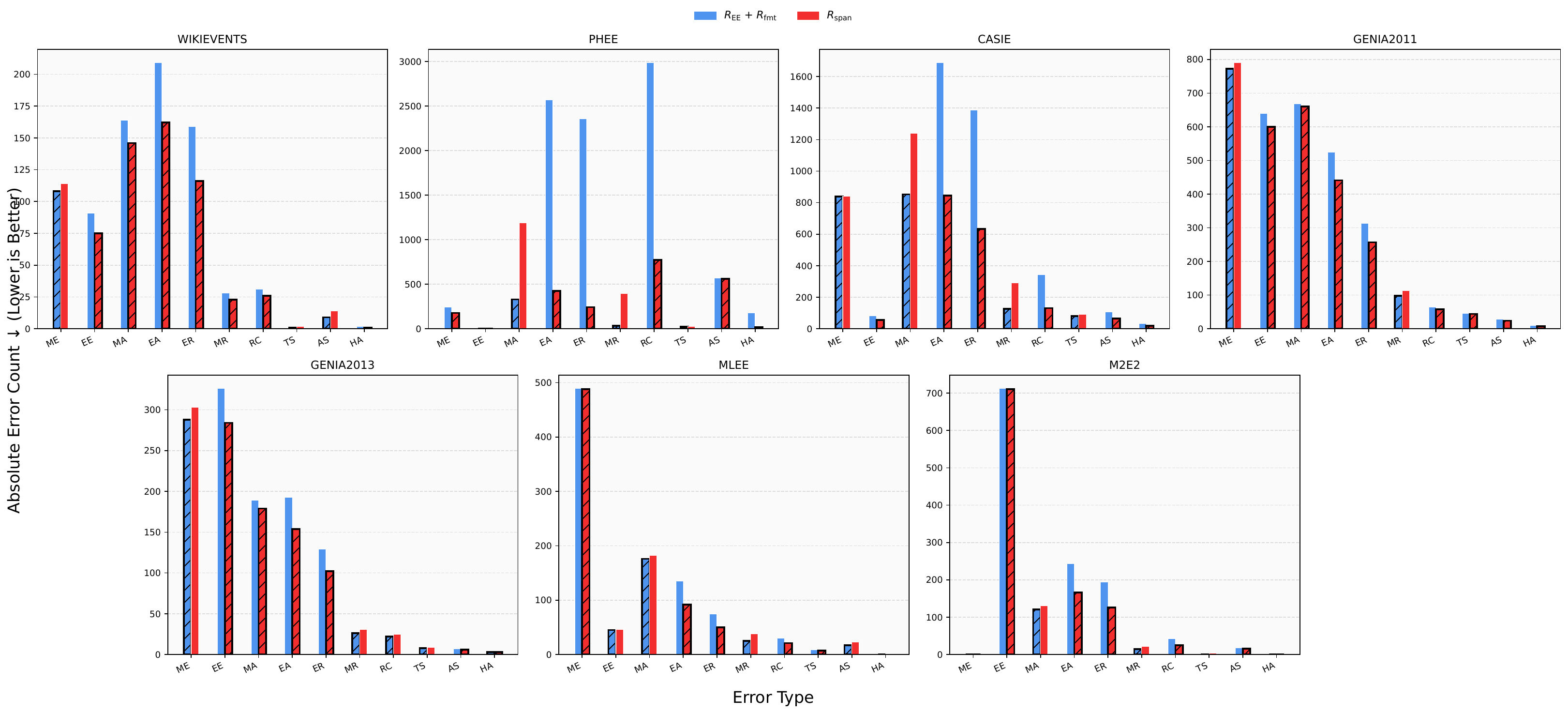}
    \caption{Comparative evaluation of error categories across datasets. $R_{\text{EE}}$ + $R_{\text{fmt}}$ Vs. +$R_{\text{span}}$ }
    \label{fig:absolute_error_comparison_span}
\vspace{-6mm}     
\end{figure*}

\begin{figure*}[ht]
    \centering
    \includegraphics[width=\textwidth]{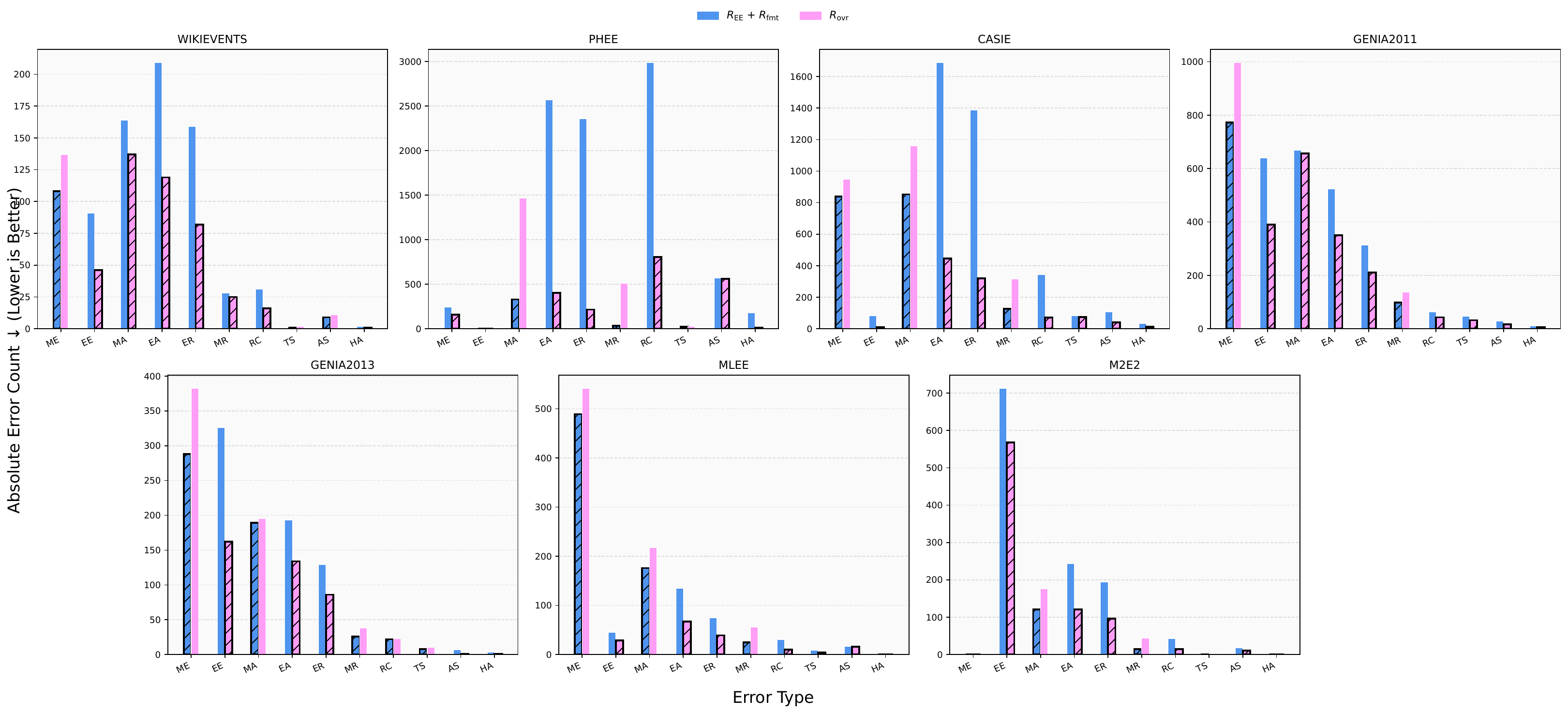}
    \caption{Comparative evaluation of error categories across datasets. $R_{\text{EE}}$ + $R_{\text{fmt}}$ Vs. +$R_{\text{ovr}}$ }
    \label{fig:absolute_error_comparison_over}
\vspace{-6mm}     
\end{figure*}

\begin{figure*}[ht]
    \centering
    \includegraphics[width=\textwidth]{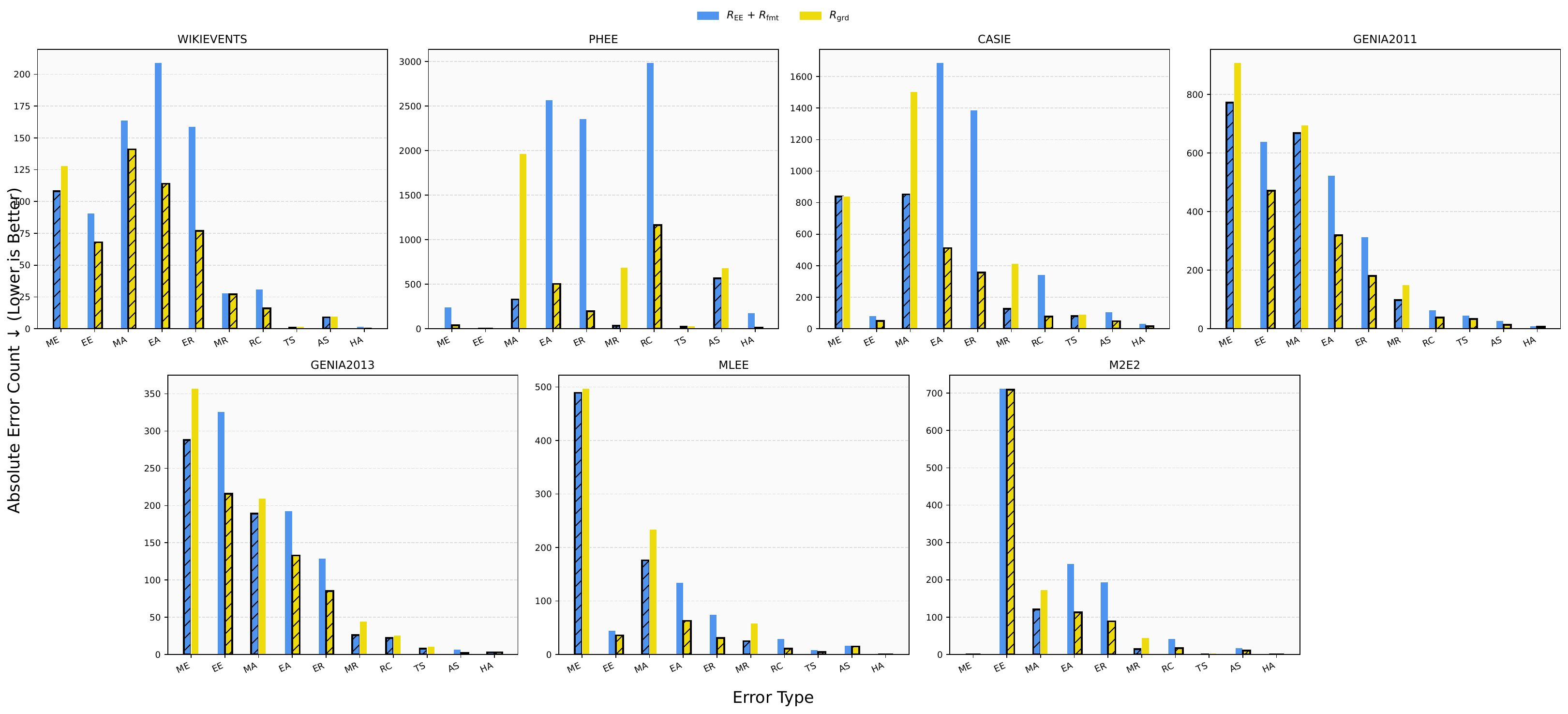}
    \caption{Comparative evaluation of error categories across datasets. $R_{\text{EE}}$ + $R_{\text{fmt}}$ Vs. +$R_{\text{grd}}$ }
    \label{fig:absolute_error_comparison_grd}
\vspace{-6mm}     
\end{figure*}

\begin{figure*}[ht]
    \centering
    \includegraphics[width=\textwidth]{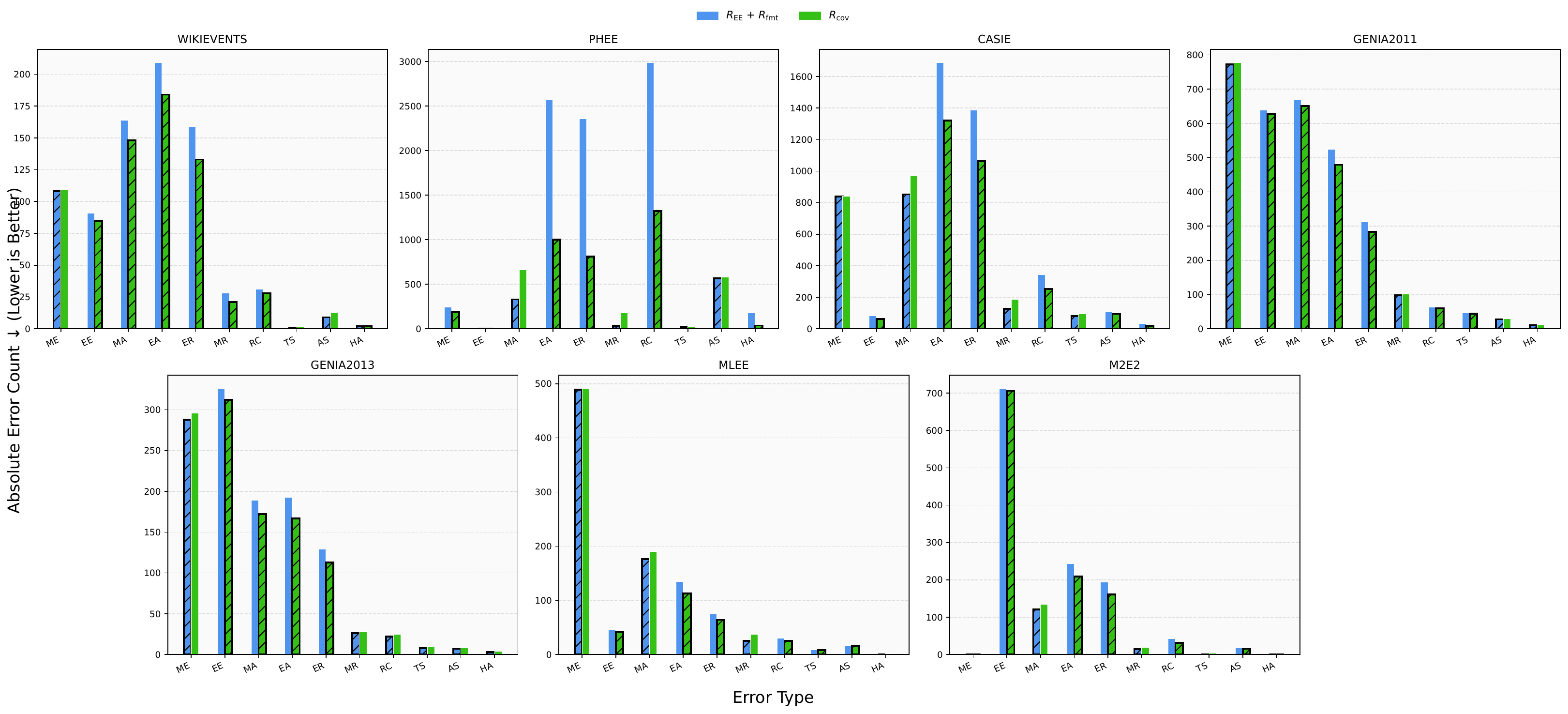}
    \caption{Comparative evaluation of error categories across datasets. $R_{\text{EE}}$ + $R_{\text{fmt}}$ Vs. +$R_{\text{cov}}$ }
    \label{fig:absolute_error_comparison_multi}
\vspace{-6mm}     
\end{figure*}

\begin{figure*}[ht]
    \centering
    \includegraphics[width=\textwidth]{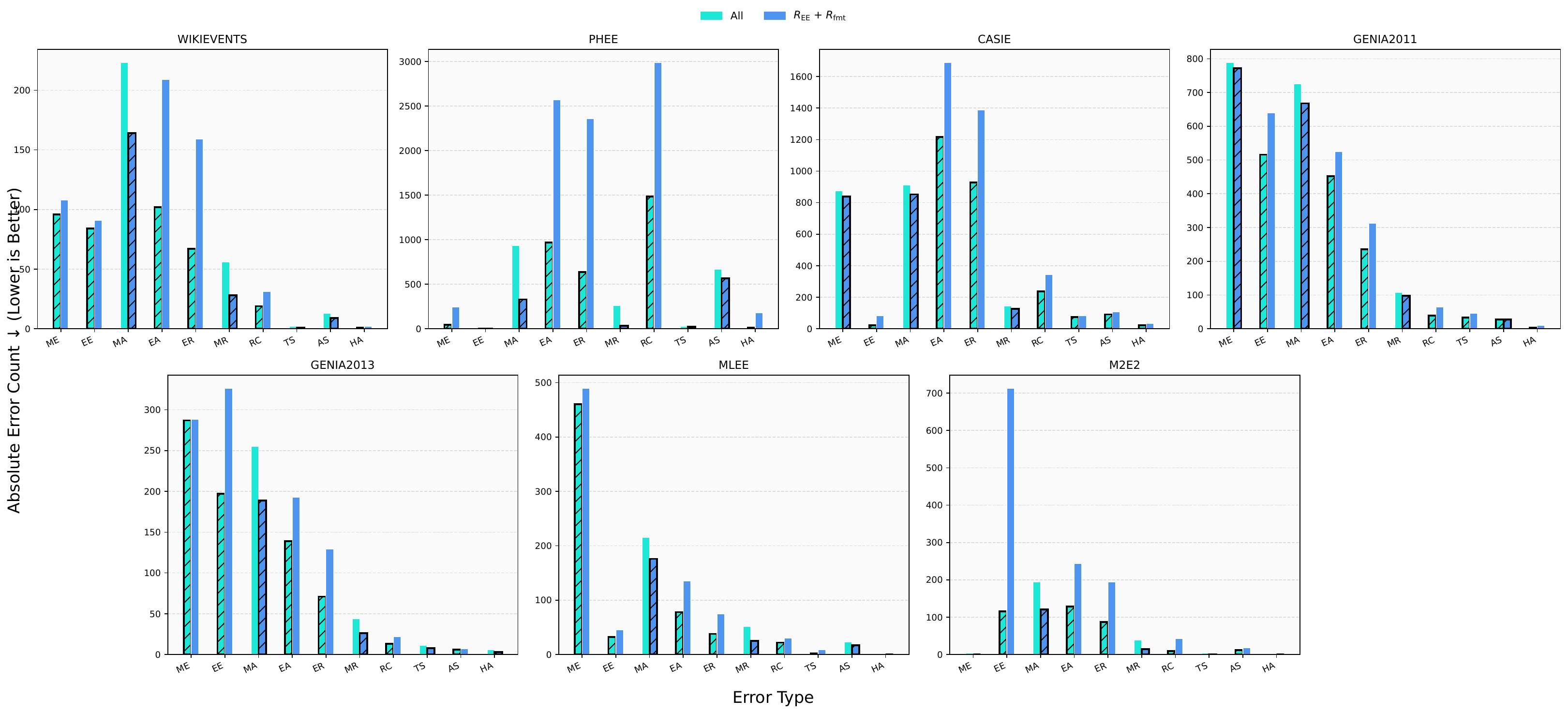}
    \caption{Comparative evaluation of error categories across datasets. $R_{\text{EE}}$ + $R_{\text{fmt}}$ Vs. All }
    \label{fig:absolute_error_comparison_all}
\vspace{-6mm}     
\end{figure*}

Figures~\ref{fig:absolute_error_comparison_span},~\ref{fig:absolute_error_comparison_over},~\ref{fig:absolute_error_comparison_grd}, ~\ref{fig:absolute_error_comparison_multi} and  ~\ref{fig:absolute_error_comparison_all} show a detailed comparative evaluation of error categories across datasets for each proposed reward. The acronyms of error categories are defined as follows: \textbf{ME}: Missing events, \textbf{EE}: Extra events, \textbf{EE}: Extra events, \textbf{MA}: Missing arguments, \textbf{EA}: Extra arguments, \textbf{RC}: Role confusion, \textbf{HA}: Hallucinations, \textbf{TS}: Trigger span boundary errors, \textbf{AS}: Argument span boundary errors, \textbf{ER}: Extra roles, \textbf{EE}: Extra events, \textbf{MR}: Missing roles. Error categories are defined in Table~\ref{tab:error_categories}.

\begin{table*}[ht]
\centering
\small
\renewcommand{\arraystretch}{1.3}
\begin{tabular}{llp{9cm}}
\toprule
\textbf{Acronym} & \textbf{Error Category} & \textbf{Definition} \\
\midrule
\multicolumn{3}{l}{\textit{Event-level Errors}} \\
\midrule
ME & Missing Events & A gold event instance is absent from the model's output; the trigger and all its associated arguments are undetected. \\
EE & Extra Events & The model predicts an event instance with no corresponding gold event; a spurious trigger is generated that is not grounded in the annotation. \\
\midrule
\multicolumn{3}{l}{\textit{Argument-level Errors}} \\
\midrule
MA & Missing Arguments & A gold argument span is not predicted for an otherwise correctly identified event; the event is detected but one or more of its argument slots are left unfilled. \\
EA & Extra Arguments & The model predicts one or more argument spans that have no corresponding gold argument; spurious fillers are generated beyond the gold annotation. \\
ER & Extra Roles & The model populates an argument role that does not exist in the gold annotation for the predicted event type, introducing schema-inconsistent role assignments. \\
MR & Missing Roles & A role defined in the gold event schema is entirely absent from the predicted event instance, resulting in incomplete role coverage. \\
RC & Role Confusion & An argument span is correctly extracted from the source text but assigned to an incorrect semantic role within the event schema; the span is right but the role label is wrong. \\
\midrule
\multicolumn{3}{l}{\textit{Span-level Errors}} \\
\midrule
TS & Trigger Span Error & The predicted trigger span does not exactly match the gold trigger boundary; includes both under-specified spans (partial overlap) and over-specified spans (superset of the gold mention). \\
AS & Argument Span Error & The predicted argument span does not exactly match the gold argument boundary; includes both partial and over-extended span predictions relative to the minimal gold span. \\
\midrule
\multicolumn{3}{l}{\textit{Faithfulness Errors}} \\
\midrule
HA & Hallucination & The predicted trigger or argument span does not appear verbatim in the source input text; the model generates mentions unsupported by the source document. \\
\midrule
\multicolumn{3}{l}{\textit{Structural Errors}} \\
\midrule
PE & Parsing Error & The model output cannot be parsed into the required code-based structured representation; the generated Python code is syntactically malformed or fails to instantiate valid event objects. \\
\bottomrule
\end{tabular}
\caption{Definitions of all error categories used in the error analysis (Section~\ref{sec:error_analysis}). 
Categories are grouped into five types: event-level errors concerning the detection of event instances, 
argument-level errors concerning role assignment and completeness, span-level errors concerning boundary 
precision, faithfulness errors concerning grounding in the source text, and structural errors concerning 
the syntactic validity of the generated output.}
\label{tab:error_categories}
\end{table*}

\begin{table*}[t]
\centering
\small
\resizebox{1\textwidth}{!}{
\setlength{\tabcolsep}{4pt}
\begin{tabular}{
l
*{7}{S[table-format=2.2]}
*{7}{S[table-format=2.2]}
*{7}{S[table-format=2.2]}
}
\toprule
\textbf{Method}
& \multicolumn{7}{c}{\textbf{WikiEvents}}
& \multicolumn{7}{c}{\textbf{PHEE}}
& \multicolumn{7}{c}{\textbf{CASIE}} \\
\cmidrule(lr){2-8} \cmidrule(lr){9-15} \cmidrule(lr){16-22}
& {TI} & {TC} & {AI} & {AC} & {AI+} & {AC+} & {\textbf{Avg.}}
& {TI} & {TC} & {AI} & {AC} & {AI+} & {AC+} & {\textbf{Avg.}}
& {TI} & {TC} & {AI} & {AC} & {AI+} & {AC+} & {\textbf{Avg.}} \\
\midrule

GPT-4o~\cite{hurst2024gpt}
& 24.58 & 20.88 & 5.43 & 4.96 & 3.22 & 2.77 & 10.31
& 2.84 & 2.84 & 2.24 & 1.82 & 1.27 & 0.98 & 2.00
& 4.58 & 4.58 & 5.12 & 4.38 & 1.15 & 0.99 & 3.47
\\

GPT-5.4-mini~\cite{singh2025openai}
& 32.04 & 23.57 & 2.28 & 1.94 & 1.98 & 1.65 & 10.58
& 50.27 & 47.03 & 5.42 & 4.46 & 3.34 & 2.66 & 18.86
& 12.86 & 12.60 & 3.01 & 2.29 & 0.85 & 0.69 & 5.38
\\

GPT-5.4~\cite{singh2025openai}
& 26.10 & 19.90 & 0.96 & 0.96 & 0.95 & 0.95 & 8.30
& 4.09 & 3.90 & 0.23 & 0.16 & 0.09 & 0.08 & 1.42 
& 13.82 & 13.82 & 3.89 & 3.12 & 1.45 & 1.23 & 6.22
\\
\midrule

Gollie-7B~\cite{sainz2024gollie}
& 30.17 & 21.32 & 9.96 & 8.62 & 6.57 & 5.28 & 13.65
& 44.15 & 42.48 & 41.73 & 17.74 & 25.30 & 11.21 & 30.44
& 10.07 & 9.40 & 11.01 & 9.23 & 2.71 & 2.42 & 7.47 \\

Gollie-13B~\cite{sainz2024gollie}
& 31.01 & 19.02 & 10.51 & 9.48 & 7.25 & 6.53 & 13.97
& 44.74 & 42.26 & 41.83 & 16.59 & 24.78 & 10.28 & 30.08
& 8.92 & 8.52 & 11.28 & 9.59 & 2.04 & 1.76 & 7.02 \\

Gollie-34B~\cite{sainz2024gollie}
& 23.94 & 18.62 & 9.80 & 7.97 & 5.85 & 4.35 & 11.75
& 36.10 & 33.41 & 41.54 & 21.35 & 20.57 & 11.57 & 27.42
& 7.96 & 7.82 & 13.26 & 11.52 & 2.15 & 1.86 & 7.43 \\

Gollie-7B SFT
& 29.26 & 21.00 & 9.75 & 8.59 & 6.30 & 4.98 & 13.31
& 43.80 & 42.47 & 41.64 & 17.74 & 25.12 & 11.20 & 30.33
& 10.46 & 9.66 & 11.12 & 9.44 & 2.70 & 2.37 & 7.62 \\

Gollie-13B SFT
& 30.66 & 18.68 & 10.31 & 9.25 & 7.25 & 6.51 & 13.78
& 44.74 & 42.26 & 41.85 & 16.59 & 24.78 & 10.28 & 30.08
& 8.92 & 8.52 & 11.28 & 9.59 & 2.04 & 1.76 & 7.02 \\

Gollie-34B SFT
& 23.94 & 18.62 & 9.80 & 7.97 & 5.85 & 4.35 & 11.75
& 36.10 & 33.41 & 41.54 & 21.35 & 20.57 & 11.57 & 27.42
& 7.96 & 7.82 & 13.26 & 11.52 & 2.15 & 1.86 & 7.43 \\


\textnormal{ADELIE}$_{\textnormal{DPO}}$ \cite{qi2024adelie}$^\dagger$
& 37.74 & 28.99 & 13.42 & 11.25 & 9.78 & 8.01 & 18.20
& 63.37 & 61.03 & 62.97 & 50.09 & 43.59 & 34.59 & 52.61
& 13.56 & 12.80 & 16.74 & 13.04 & 5.19 & 4.09 & 10.90 \\

EventRL~\cite{gao2024eventrl}$^\dagger$
& 39.88 & 32.75 & 15.14 & 12.39 & 11.70 & 9.40 & 20.21
& 67.92 & 66.80 & 62.29 & 46.06 & 44.95 & 33.54 & 53.59
& 20.46 & 19.76 & 23.66 & 17.10 & 7.64 & 5.46 & 15.68 \\


\midrule
EAGER
& 38.34 & 30.01 & 16.60 & 14.69 & 12.54 & 10.89 & 20.51
& 68.94 & 68.53 & 67.73 & 58.96 & 49.40 & 43.26 & \bfseries 59.47
& 19.92 & 18.72 & 22.54 & 17.48 & 6.71 & 5.29 & 15.11 \\

\bottomrule
\end{tabular}
}
\caption{Full results on WikiEvents, PHEE and CASIE datasets.}
\label{tab:full_results_WikiEvents_PHEE_CASIE}
\end{table*}

\begin{table*}[t]
\centering
\small
\resizebox{1\textwidth}{!}{
\setlength{\tabcolsep}{4pt}
\begin{tabular}{
l
*{7}{S[table-format=2.2]}
*{7}{S[table-format=2.2]}
}
\toprule
\textbf{Method}
& \multicolumn{7}{c}{\textbf{Genia2011}}
& \multicolumn{7}{c}{\textbf{Genia2013}} \\
\cmidrule(lr){2-8} \cmidrule(lr){9-15}
& {TI} & {TC} & {AI} & {AC} & {AI+} & {AC+} & {\textbf{Avg.}}
& {TI} & {TC} & {AI} & {AC} & {AI+} & {AC+} & {\textbf{Avg.}} \\
\midrule

GPT-4o~\cite{hurst2024gpt}
& 27.67 & 24.55 & 8.23 & 7.35 & 3.52 & 3.28 & 12.43
& 24.92 & 22.01 & 7.53 & 5.98 & 2.46 & 2.32 & 10.87 \\

GPT-5.4-mini~\cite{singh2025openai}
& 34.85 & 30.24 & 2.38 & 2.01 & 1.15 & 1.06 & 11.95
& 30.07 & 27.01 & 2.06 & 1.43 & 1.61 & 1.20 & 10.56 \\

GPT-5.4~\cite{singh2025openai}
& 31.90 & 29.89 & 3.37 & 2.66 & 2.01 & 1.83 & 11.94
& 21.82 & 19.39 & 3.41 & 3.01 & 2.58 & 2.21 & 8.74 \\

\midrule

Gollie-7B~\cite{sainz2024gollie}
& 27.84 & 19.41 & 11.08 & 8.61 & 4.66 & 3.88 & 12.58
& 20.15 & 14.56 & 10.69 & 6.24 & 4.02 & 2.46 & 9.69 \\

Gollie-13B~\cite{sainz2024gollie}
& 27.44 & 12.70 & 8.74 & 6.71 & 3.92 & 2.44 & 10.33
& 25.20 & 12.14 & 9.01 & 6.13 & 2.91 & 1.99 & 9.56 \\

Gollie-34B~\cite{sainz2024gollie}
& 27.87 & 16.45 & 10.02 & 6.44 & 4.14 & 2.64 & 11.26
& 23.13 & 11.38 & 6.98 & 2.14 & 3.16 & 1.34 & 8.02 \\

Gollie-7B SFT
& 27.95 & 19.46 & 10.44 & 7.78 & 4.50 & 3.73 & 12.31
& 21.24 & 15.10 & 11.64 & 6.99 & 4.32 & 2.75 & 10.34 \\

Gollie-13B SFT
& 27.44 & 12.70 & 8.74 & 6.71 & 3.92 & 2.44 & 10.33
& 25.20 & 12.14 & 9.01 & 6.13 & 2.91 & 1.99 & 9.56 \\

Gollie-34B SFT
& 27.87 & 16.45 & 10.02 & 6.44 & 4.14 & 2.64 & 11.26
& 23.13 & 11.38 & 6.98 & 2.14 & 3.16 & 1.34 & 8.02 \\


\textnormal{ADELIE}$_{\textnormal{DPO}}$ \cite{qi2024adelie}$^\dagger$
& 39.06 & 29.45 & 20.43 & 17.65 & 10.06 & 8.79 & 20.91
& 39.05 & 31.72 & 16.92 & 13.25 & 9.09 & 6.98 & 19.50 \\

EventRL~\cite{gao2024eventrl}$^\dagger$
& 43.46 & 38.80 & 25.19 & 23.18 & 13.57 & 12.64 & 26.14
& 43.58 & 38.76 & 20.50 & 18.69 & 12.04 & 11.25 & 24.14 \\


\midrule
EAGER
& 43.77 & 39.15 & 23.54 & 21.98 & 12.56 & 12.06 & 25.51
& 44.14 & 39.67 & 21.23 & 19.71 & 10.28 & 9.92 & 24.16 \\

\bottomrule
\end{tabular}
}
\caption{Full results on Genia2011 and Genia2013 datasets.}
\label{tab:full_results_genia11_13}
\end{table*}

\begin{table*}[t]
\centering
\small
\resizebox{1\textwidth}{!}{
\setlength{\tabcolsep}{4pt}
\begin{tabular}{
l
*{7}{S[table-format=2.2]}
*{7}{S[table-format=2.2]}
}
\toprule
\textbf{Method}
& \multicolumn{7}{c}{\textbf{MLEE}}
& \multicolumn{7}{c}{\textbf{M2E2}} \\
\cmidrule(lr){2-8} \cmidrule(lr){9-15}
& {TI} & {TC} & {AI} & {AC} & {AI+} & {AC+} & {\textbf{Avg.}}
& {TI} & {TC} & {AI} & {AC} & {AI+} & {AC+} & {\textbf{Avg.}} \\
\midrule

GPT-4o~\cite{hurst2024gpt}
& 28.02 & 24.49 & 8.79 & 7.57 & 5.99 & 5.23 & 13.35
& 20.78 & 19.91 & 8.67 & 8.56 & 6.97 & 6.91 & 11.97 \\

GPT-5.4-mini~\cite{singh2025openai}
& 32.60 & 25.50 & 2.56 & 2.55 & 0.49 & 0.49 & 10.70
& 35.16 & 33.59 & 4.26 & 3.58 & 3.61 & 2.97 & 13.86 \\

GPT-5.4~\cite{singh2025openai}
& 34.59 & 30.40 & 2.09 & 1.56 & 1.51 & 1.26 & 11.90
& 33.96 & 33.21 & 11.52 & 10.77 & 8.83 & 8.16 & 17.74 \\

\midrule

Gollie-7B~\cite{sainz2024gollie}
& 25.35 & 13.04 & 5.14 & 3.07 & 2.57 & 1.78 & 8.49
& 51.16 & 47.84 & 27.88 & 24.81 & 21.69 & 19.45 & 32.14 \\

Gollie-13B~\cite{sainz2024gollie}
& 25.86 & 17.99 & 9.01 & 6.61 & 5.41 & 3.90 & 11.46
& 61.63 & 59.82 & 33.56 & 29.69 & 25.98 & 23.05 & 38.96 \\

Gollie-34B~\cite{sainz2024gollie}
& 18.83 & 14.25 & 10.55 & 7.16 & 3.76 & 2.94 & 9.58
& 50.17 & 48.18 & 30.06 & 26.59 & 22.30 & 19.49 & 32.80 \\

Gollie-7B + SFT
& 26.27 & 13.21 & 5.77 & 3.68 & 2.78 & 1.98 & 8.95
& 52.17 & 48.83 & 28.41 & 25.33 & 21.85 & 19.59 & 32.70 \\

Gollie-13B + SFT
& 25.86 & 17.99 & 9.01 & 6.61 & 5.41 & 3.90 & 11.46
& 61.63 & 59.82 & 33.56 & 29.69 & 25.98 & 23.05 & 38.96 \\

Gollie-34B + SFT
& 18.83 & 14.25 & 10.55 & 7.16 & 3.76 & 2.94 & 9.58
& 50.17 & 48.18 & 30.06 & 26.59 & 22.30 & 19.49 & 32.80 \\


\textnormal{ADELIE}$_{\textnormal{DPO}}$ \cite{qi2024adelie}$^\dagger$
& 39.74 & 27.15 & 12.20 & 8.17 & 8.57 & 5.54 & 16.89
& 65.26 & 61.50 & 34.34 & 30.62 & 26.98 & 24.26 & 40.49 \\

EventRL~\cite{gao2024eventrl}$^\dagger$
& 43.96 & 30.91 & 18.43 & 16.33 & 11.56 & 10.24 & 21.90
& 56.45 & 50.32 & 24.16 & 17.49 & 18.74 & 13.77 & 30.16 \\


\midrule

EAGER
& 47.85 & 37.08 & 19.28 & 16.71 & 13.22 & 11.12 & \bfseries 24.21
& 61.57 & 58.12 & 38.46 & 33.63 & 29.46 & 26.74 & \bfseries 41.33 \\

\bottomrule
\end{tabular}
}
\caption{Full results on MLEE and M2E2 datasets.}
\label{tab:full_results_mlee_m2e2}
\end{table*}


\section{Examples Event Schema with Generated Annotation Guidelines}
Figures~\ref{fig:WikiEvents_event_schema_exp},~\ref{fig:PHEE_event_schema_exp},~\ref{fig:CASIE_event_schema_exp},~\ref{fig:Genia2011_event_schema_exp},~\ref{fig:Genia2013_event_schema_exp},~\ref{fig:MLEE_event_schema_exp} and ~\ref{fig:M2E2_event_schema_exp}, illustrate  examples of one event scheme augmented with annotation guidelines for each dataset.

\begin{figure*}[t]
\begin{lstlisting}[style=pythonstyle]
@dataclass
class ArtifactExistence_DamageDestroyDisableDismantle_Damage:
    """The event type ArtifactExistence_DamageDestroyDisableDismantle_Damage is triggered by the mention of damage or destruction of an artifact, which can be a physical object, structure, or entity. It involves the concept of causing harm or damage to something, resulting in its degradation or loss of functionality. Unlike ArtifactExistence_DamageDestroyDisableDismantle_Unspecified, this event type specifically focuses on damage or destruction, excluding other forms of disablement or dismantling. Triggers such as 'damage', 'destroy', 'disable', or 'dismantle' are indicative of this event type. Examples include: 'The building was damaged in the earthquake.', 'The car was destroyed in the accident.', 'The bridge was disabled by the protesters.', 'The old factory was dismantled for redevelopment.'"""
    mention: str  # The mention argument refers to the verb or phrase that indicates the occurrence of the event. Examples are: 'damage', 'destroy', 'disable', 'dismantle', 'harm', 'destroyed', 'damaged', 'disabled', or 'dismantled'. Variations include 'caused damage', 'caused destruction', 'caused harm', or'resulted in damage'.
    Artifact: List  # The artifact argument refers to the object, structure, or entity that is damaged, destroyed, disabled, or dismantled. Examples are: 'building', 'car', 'bridge', 'factory', 'property', or 'equipment'.
    DamagerDestroyer: List
    Instrument: List  # The instrument argument refers to the tool, device, or means used to cause the damage or destruction. Examples are: 'bomb', 'grenades', 'fire', or 'chemicals'.
    Place: List  # The place argument refers to the location where the event occurs. Examples are: 'city', 'building', 'park', or'street'.
    Damager: List  # The damager argument refers to the entity responsible for causing the damage or destruction. Examples are: 'terrorist group', 'vandal', 'accident', 'natural disaster', or 'human error'.
\end{lstlisting}
\caption{\textbf{WikiEvents} event schema python class example}
\label{fig:WikiEvents_event_schema_exp}
\end{figure*}

\begin{figure*}[t]
\begin{lstlisting}[style=pythonstyle]
@dataclass
class Adverse_event:
    """The event is triggered by the mention of an adverse event, which is a harmful or undesirable effect that occurs as a result of a treatment, medication, or other intervention. Adverse events can be caused by a combination of drugs, a single drug, or a treatment duration. Examples are intravenous azithromycin-induced ototoxicity, unaccountable severe hypercalcemia in a patient treated for hypoparathyroidism with dihydrotachysterol, and prolonged severe 5-fluorouracil-associated neurotoxicity in a patient with dihydropyrimidine dehydrogenase deficiency. Unlike other events, adverse events are not therapeutic or beneficial, and they are often characterized by negative outcomes such as hematologic adverse reactions, pulmonary toxicity, or supravenous hyperpigmentation. Triggers such as 'developed', 'induced', 'become', 'on', and 'by' are indicative of adverse events, not other event types."""
    mention: str  # The mention argument indicates the type of relationship between the subject and the adverse event. Examples are 'developed', 'induced', 'become', 'on', and 'by'.
    Combination_Drug: List  # The Combination_Drug argument lists the medications that were combined to cause the adverse event. Examples are'methotrexate' and 'bleomycin', 'thionamide', and 'warfarin'.
    Effect: List  # The Effect argument describes the specific adverse effect that occurred. Examples are 'acute digital ischemia and gangrene', 'hematologic adverse reactions', and 'pulmonary toxicity'.
    Subject: List  # The Subject argument identifies the person or entity that experienced the adverse event. Examples are 'a 57-year-old man', 'a patient', and 'a CD30 (Ki-1)-positive anaplastic large-cell lymphoma'.
    Subject_Age: List  # The Subject_Age argument specifies the age of the subject. Examples are '57-year-old', '44-year-old', and '20-year-old'.
    Subject_Disorder: List  # The Subject_Disorder argument describes any pre-existing medical condition the subject had before the adverse event. Examples are 'acral erythrocyanosis', 'hypoparathyroidism', and 'rheumatoid arthritis'.
    Subject_Gender: List  # The Subject_Gender argument specifies the gender of the subject. Examples are'man', 'woman', and 'patient'.
    Subject_Population: List  # The Subject_Population argument identifies the population the subject belongs to. Examples are 'over 75s', and 'patients'.
    Subject_Race: List  # The Subject_Race argument specifies the race of the subject. 
    Treatment: List  # The Treatment argument lists the treatment or medication that caused the adverse event. Examples are 'combined chemotherapy (bleomycin and methotrexate)', 'thionamide', and 'CHOP chemotherapy'.
    Treatment_Disorder: List  # The Treatment_Disorder argument describes the medical condition being treated. Examples are'metastatic squamous cell carcinoma of the hypopharynx', 'hypoparathyroidism', and 'a CD30 (Ki-1)-positive anaplastic large-cell lymphoma'.
    Treatment_Dosage: List  # The Treatment_Dosage argument specifies the dosage of the treatment. Examples are [], 'chronic', and 'therapeutic doses'.
    Treatment_Drug: List  # The Treatment_Drug argument lists the specific medication used in the treatment. Examples are'methotrexate', 'thionamide', and 'CHOP'.
    Treatment_Duration: List  # The Treatment_Duration argument specifies the duration of the treatment. 
    Treatment_Freq: List  # The Treatment_Freq argument specifies the frequency of the treatment.
    Treatment_Route: List  # The Treatment_Route argument specifies the route of administration of the treatment. Examples  'chemotherapy'.
    Treatment_Time_elapsed: List  # The Treatment_Time_elapsed argument specifies the time elapsed since the treatment started.
\end{lstlisting}
\caption{\textbf{PHEE} event schema python class example}
\label{fig:PHEE_event_schema_exp}
\end{figure*}

\begin{figure*}[t]
\begin{lstlisting}[style=pythonstyle]
@dataclass
class Attack_Ransom:
    """The event is triggered by the occurrence of a ransomware attack, where an attacker demands payment in exchange for restoring access to encrypted data. The event is characterized by the use of ransomware, encryption of data, and the demand for payment. Unlike other types of attacks, Attack_Ransom events involve the use of ransomware and the demand for payment to restore access to data. Triggers such as 'demanded a ransom' and 'ransomware attacks' are indicative of Attack_Ransom events. Examples of Attack_Ransom events include instances where data is encrypted and a ransom is demanded, such as 'demanded in payment' and 'demanded a ransom'."""
    mention: str  # The mention argument refers to the specific mention of the event, which can be a phrase or sentence that describes the event. Examples are 'demanded in payment', 'demanded a ransom', 'ransomware attacks', 'ransom', and 'Paying ransomware'.
    Attack_Pattern: List  # The Attack_Pattern argument refers to the specific pattern of behavior exhibited by the attacker. Examples are 'threatened to delete the files' and'remotely wipe millions of iPhones and iCloud accounts'.
    Attacker: List  # The Attacker argument refers to the identity or description of the individual or group responsible for the attack. Examples are 'cyber fraudsters', 'one lonesome individual', and 'criminals'.
    Damage_Amount: List  # The Damage_Amount argument refers to the estimated or reported amount of damage caused by the attack. Examples are empty lists, indicating that the amount of damage is unknown or not reported.
    Payment_Method: List  # The Payment_Method argument refers to the method by which the attacker demands or receives payment. Examples are empty lists, indicating that the payment method is unknown or not reported.
    Place: List  # The Place argument refers to the location where the attack occurred. Examples are empty lists, indicating that the location is unknown or not reported.
    Price: List  # The Price argument refers to the amount of money demanded or paid as a ransom. Examples are 'USD 216', 'USD 2,000', and empty lists, indicating that the price is unknown or not reported.
    Time: List  # The Time argument refers to the date or time when the attack occurred. Examples are '96 hours', 'April 7', and empty lists, indicating that the time is unknown or not reported.
    Tool: List  # The Tool argument refers to the specific tool or malware used by the attacker. Examples are'malware', 'WannaCry', 'Bad Rabbit', and empty lists, indicating that the tool is unknown or not reported.
    Victim: List  # The Victim argument refers to the individual or organization affected by the attack. Examples are 'Texan city', 'equipment', 'Apple', 'other businesses', and 'South Koreans'.
\end{lstlisting}
\caption{\textbf{CASIE} event schema python class example}
\label{fig:CASIE_event_schema_exp}
\end{figure*}

\begin{figure*}[t]
\begin{lstlisting}[style=pythonstyle]
@dataclass
class Protein_catabolism:
    """The event is triggered by the process of breaking down or degrading proteins. This process can be induced by various mechanisms, including proteasomal degradation, ubiquitination, and phosphorylation. The key characteristics of this event include the degradation of proteins, which can be a result of various cellular processes. Unlike other events such as Gene_expression, this event is focused on the breakdown of proteins rather than their synthesis. Triggers such as 'degradation', 'proteolytically degraded', and 'ubiquitination' are indicative of this event type. Examples of protein degradation include the breakdown of IkappaBalpha, A3G, and p27kip1. The scope of this event is limited to the degradation of proteins, and it does not include other cellular processes such as transcription or phosphorylation."""
    mention: str  # Examples are 'degradation', 'proteolytically degraded', 'ubiquitination', and other terms indicating protein breakdown. The role of the mention argument is to specify the type of protein degradation event. Importance: The mention argument is crucial in identifying the type of protein degradation event, which is essential in understanding the cellular process.
    Theme: List  # Examples are 'IkappaBalpha', 'A3G', 'p105', and other protein names. The role of the Theme argument is to specify the protein being degraded. Importance: The Theme argument is essential in identifying the protein undergoing degradation, which is critical in understanding the cellular process.
\end{lstlisting}
\caption{\textbf{Genia2011} event schema python class example}
\label{fig:Genia2011_event_schema_exp}
\end{figure*}

\begin{figure*}[t]
\begin{lstlisting}[style=pythonstyle]
@dataclass
class Phosphorylation:
    """The event is triggered by the phosphorylation of a molecule, typically a protein. Phosphorylation is a chemical reaction that adds a phosphate group to a molecule, often altering its function or activity. This event is characterized by the transfer of a phosphate group from a phosphate donor to a protein, resulting in a change in the protein's conformation or activity. Examples of triggers include 'phosphorylated', 'phosphorylate', and 'phosphorylation'. Unlike Protein_catabolism, this event does not involve the degradation of a protein, and unlike Protein_modification, it does not involve the addition or removal of a non-phosphate group. Triggers such as 'phosphorylation' are indicative of Phosphorylation, not Protein_modification."""
    mention: str  # Examples are 'phosphorylated', 'phosphorylate', and 'phosphorylation'. The mention argument represents the trigger or indicator of the phosphorylation event. It is the key term that indicates that phosphorylation has occurred. This argument is crucial in identifying the event and distinguishing it from other types of protein modifications.
    Cause: List  # Examples are empty lists. The Cause argument represents the stimulus or trigger that leads to the phosphorylation event. It is the factor that initiates the phosphorylation reaction. In some cases, the cause may be a specific molecule, such as a protein or a ligand, that binds to the target protein and triggers the phosphorylation.
    Site: List  # Examples are empty lists. The Site argument represents the location on the protein where the phosphorylation occurs. It is the specific residue or region of the protein that is modified by the addition of a phosphate group.
    Theme: List  # Examples are '8', 'Smad1', 'RPS3', 'HSP27', 'IkappaBalpha', and 'Flag-RPS3'. The Theme argument represents the protein or molecule that is being phosphorylated. It is the target of the phosphorylation event, and its modification results in a change in its function or activity.
\end{lstlisting}
\caption{\textbf{Genia2013} event schema python class example}
\label{fig:Genia2013_event_schema_exp}
\end{figure*}

\begin{figure*}[t]
\begin{lstlisting}[style=pythonstyle]
@dataclass
class Dephosphorylation:
    """The event is triggered by the removal of a phosphate group from a protein, typically as a result of the action of a phosphatase enzyme. The event is characterized by the dephosphorylation of a specific protein site. Unlike Phosphorylation, this event does not involve the addition of a phosphate group. Triggers such as 'dephosphorylation', 'phosphatase', or 'dephosphorylated' are indicative of Dephosphorylation, not Phosphorylation. Examples are 'dephosphorylation of Mcl-1', 'phosphatase activity','removal of phosphate group'."""
    mention: str  # The event trigger, which is the dephosphorylation process itself, or a related term such as 'phosphatase' or 'dephosphorylated'. Examples are 'dephosphorylation', 'phosphatase', 'dephosphorylated'.
    Site: List  # The specific location on the protein where the dephosphorylation occurs. Examples are 'Mcl-1','serine 157', 'tyrosine 123'.
    Theme: List  # The protein that undergoes dephosphorylation. Examples are 'Mcl-1', 'ERK kinase', 'histone H3'.
\end{lstlisting}
\caption{\textbf{MLEE} event schema python class example}
\label{fig:MLEE_event_schema_exp}
\end{figure*}

\begin{figure*}[t]
\begin{lstlisting}[style=pythonstyle]
@dataclass
class Justice_arrestJail:
    """The event is triggered by the arrest or detention of a person or persons, which can be initiated by authorities or law enforcement. This event type covers various contexts, including but not limited to, arrests, detentions, and imprisonments. Unlike other event types, such as Conflict_Attack, this event does not involve physical harm or violence. Triggers such as 'arrested', 'jailed', 'detained', and 'imprisoned' are indicative of this event type. Examples include 'He's been arrested and details will soon be released', 'Local media circulated a photo of what they described as the moment he was arrested', and 'A web search found reports going back five years where authorities had arrested or charged teenagers for recruiting other teen girls for prostitution'."""
    mention: str  # The mention argument represents the trigger or indicator of the event, such as 'arrested', 'jailed', 'detained', or 'imprisoned'. Examples are 'arrested', 'jailed', 'detained', 'imprisoned', 'held', 'arrests', and 'detained'.
    Agent: List  # The Agent argument represents the entity or entities responsible for the arrest or detention, which can be authorities, law enforcement, or other entities. Examples are 'Interior Ministry', 'police officers', 'authorities', and 'law enforcement'.
    Person: List  # The Person argument represents the individual or individuals arrested or detained. Examples are'sailors', 'Hendry', 'teenagers', 'Italian', 'Mahmoud Abd al-Aziz al-Mujahid', 'Rajab', and 'Daniel Ramirez Medina'.
    Place: List  # The Place argument represents the location where the arrest or detention occurred. Examples are 'Farsi Island', 'Guantanamo Bay', 'Moscow', 'Bahrain', 'Russia', 'Crimea', and 'Washington'.
\end{lstlisting}
\caption{\textbf{M2E2} event schema python class example}
\label{fig:M2E2_event_schema_exp}
\end{figure*}


\end{document}